\documentclass{article}

\usepackage[preprint, nonatbib]{neurips_2024}

\usepackage{amsmath,amsfonts,bm}

\def\eqref#1{equation~\ref{#1}}

\def\1{\bm{1}}

\DeclareMathAlphabet{\mathsfit}{\encodingdefault}{\sfdefault}{m}{sl}
\SetMathAlphabet{\mathsfit}{bold}{\encodingdefault}{\sfdefault}{bx}{n}

\def\0{\textbf{0}}
\def\1{\textbf{1}}

\def\e{\boldsymbol{e}}

\def\x{\boldsymbol{x}}

\def\X{\boldsymbol{X}}

\def\cS{\mathcal{S}}

\renewcommand{\mathbf}{\boldsymbol}

\renewcommand{\Re}{{\mathbb R}}

\usepackage[utf8]{inputenc} 
\usepackage[T1]{fontenc}    
\usepackage{hyperref}       
\usepackage{url}            
\usepackage{booktabs}       
\usepackage{amsfonts}       
\usepackage{nicefrac}       
\usepackage{microtype}      
\usepackage{xcolor}         
\usepackage{bm}
\usepackage{float}
\usepackage[
backend=biber,
style=numeric,
sorting=ynt
]{biblatex}
\usepackage{subfigure}
\usepackage{subcaption}
\usepackage{graphicx}
\usepackage{booktabs}
\usepackage{caption}
\usepackage{multirow}
\usepackage{wrapfig}
\title{Ctrl123: Consistent Novel View Synthesis via Closed-Loop Transcription}

\author{%
  Hongxiang Zhao\thanks{Equal Contribution.$^{1}$Beijing University of Posts and Telecommunications.$^{2}$Hong Kong University of Science and Technology (Guangzhou).$^{3}$International Digital Economy Academy (IDEA).$^{4}$New York University.$^{5}$Tsinghua-Berkeley Shenzhen Institute (TBSI).$^{6}$University of California, Berkeley.}$^{\;\;,1}$ \And
  Xili Dai$^{*,2,3}$ \And
  Jianan Wang$^3$ \And 
  Shengbang Tong$^4$ \AND
  Jingyuan Zhang$^5$ \And
  Weida Wang$^5$ \And
  Lei Zhang$^3$ \And
  Yi Ma$^6$
}

\newcommand{\ours}{Ctrl123}
\definecolor{darkgreen}{rgb}{0.0, 0.5, 0.0}
\newcommand{\plusvalue}[1]{\hspace{0.3em}\textcolor{darkgreen}{(+#1)}}
\begin{document}
\newcommand{\hx}[1]{{\color{blue} \textbf{#1}}}
\newcommand{\xili}[1]{{\color{red} \textbf{#1}}}

\maketitle

\begin{center}
    \centering
    \includegraphics[width=0.85\linewidth]{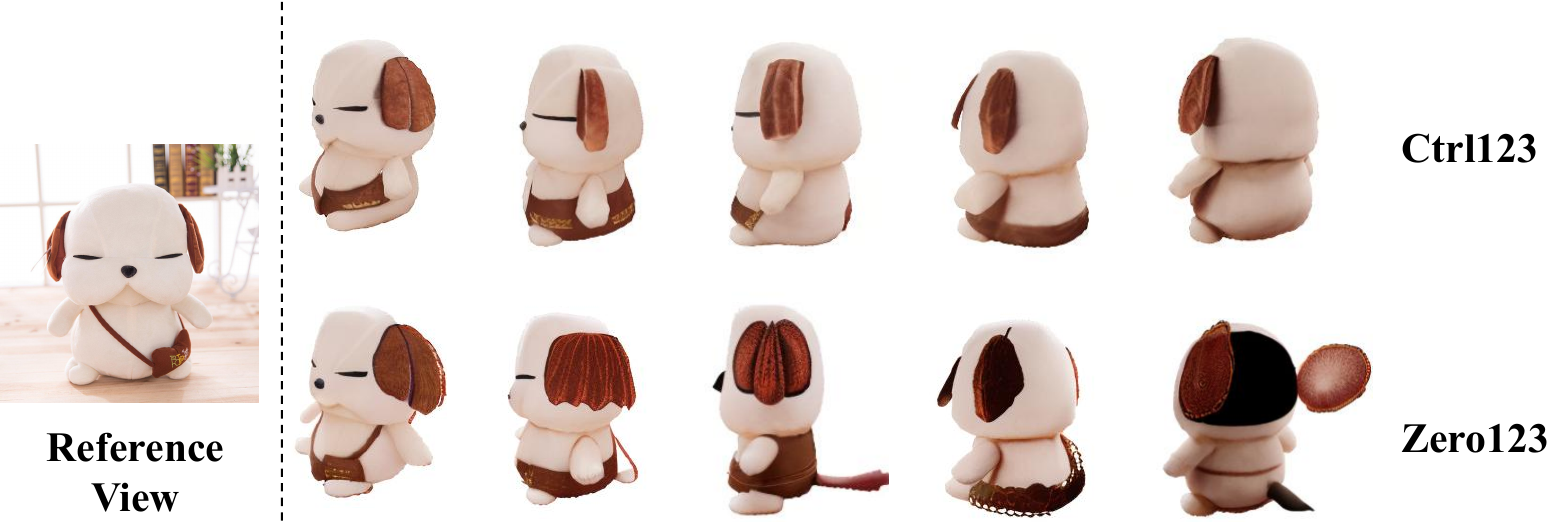}
    \captionof{figure}{\textit{\ours{}} generates more pose and appearance consistent novel views from a single image of an arbitrary object.}
    \label{fig:teaser}
    \vspace{-0.3cm}
\end{center}

\begin{abstract}
  Large image diffusion models have demonstrated zero-shot capability in novel view synthesis (NVS). However, existing diffusion-based NVS methods struggle to generate novel views that are accurately consistent with the corresponding ground truth poses and appearances, even on the training set. This consequently limits the performance of downstream tasks, such as image-to-multiview generation and 3D reconstruction. We realize that such inconsistency is largely due to the fact that it is difficult to enforce accurate pose and appearance alignment directly in the diffusion training, as mostly done by existing methods such as Zero123 \cite{liu2023zero}
  . To remedy this problem, we propose \ours{}, a {\em closed-loop} transcription-based NVS diffusion method that enforces alignment between the generated view and ground truth in a pose-sensitive feature space. Our extensive experiments demonstrate the effectiveness of \ours{} on the tasks of NVS and 3D reconstruction, achieving significant improvements in both multiview-consistency and pose-consistency over existing methods.
\end{abstract}

\vspace{-0.2cm}
\section{Introduction}
\label{sec:intro}

Recent advancements in novel view synthesis (NVS) from a single image have generated significant enthusiasm within the 3D community~\cite{poole2022dreamfusion,wang2023prolificdreamer}. NVS is one of the fundamental tasks for any 3D content generation system. It is important for generating a full 3D model from a single image captured by users or generated by text-to-image models. More recently, Zero123~\cite{liu2023zero} enhances the generalization capability
by leveraging powerful pre-trained image generation models such as Stable Diffusion (SD) ~\cite{latentdiffusion}. It fine-tunes Stable Diffusion on image renderings derived from a diverse 3D dataset~\cite{objaverse} to learn to generate a novel view conditioned on the reference view and a relative camera transformation. The zero-shot and open-world capabilities of Zero123 \cite{liu2023zero} represent a significant advancement in the field of NVS. Following the pioneering work of Zero123 \cite{liu2023zero}, subsequent studies~\cite{shi2023zero123++,song2023consistency,liu2023syncdreamer,TOSS} further advanced the overall NVS performance. Zero123-XL \cite{deitke2023objaverse} focuses on data quantity, scaling the training data to up to 10 times more; Zero123++~\cite{shi2023zero123++} restricts novel view generation from arbitrary view to 6 predetermined views, which significantly reduces the complexity of the setting of the task. 

\begin{figure}
    \centering
    \subfigure[Zero123-XL \cite{deitke2023objaverse} fails to generate consistent results in terms of pose and appearance.]{
        \begin{minipage}{0.45\linewidth}
            \centering
            \includegraphics[width=0.9\linewidth]{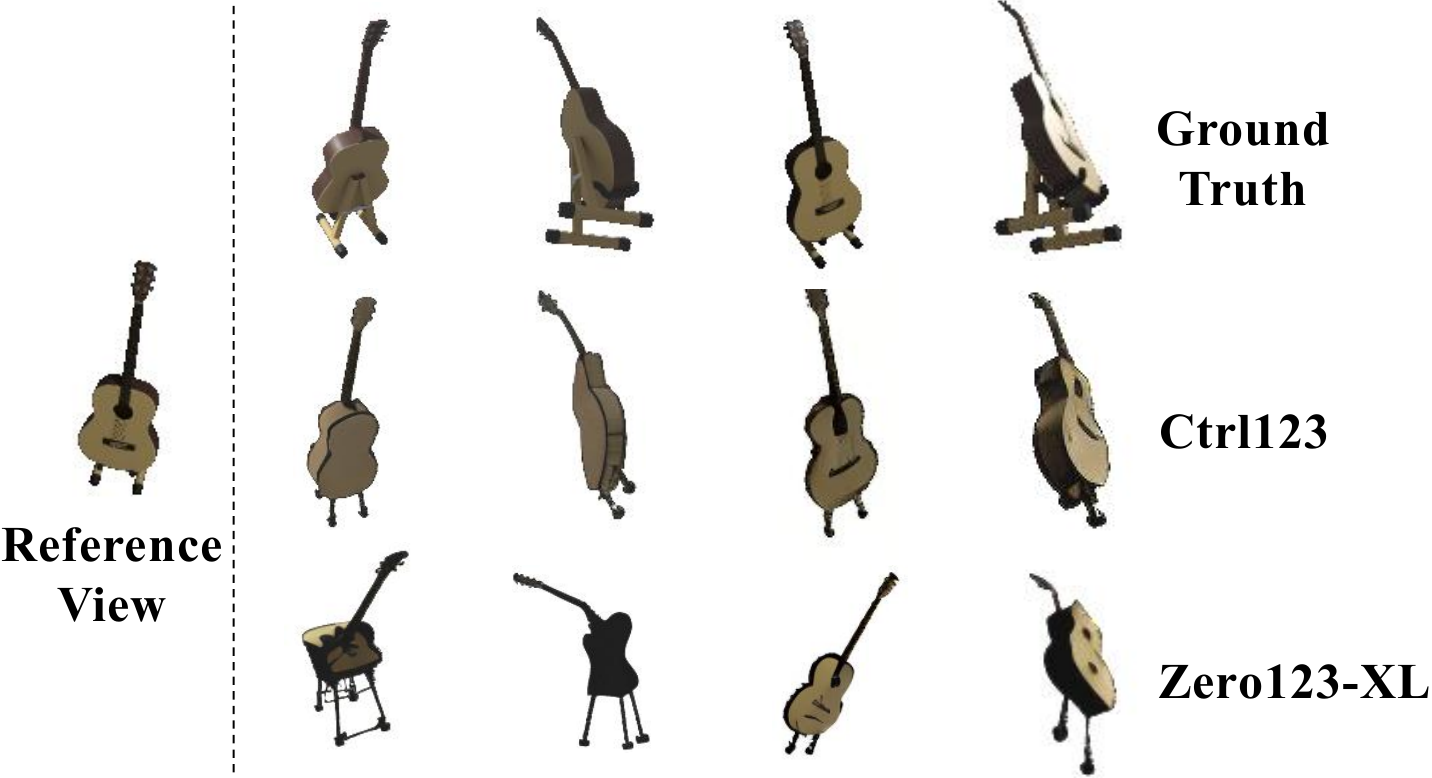}\label{fig:fail-case-a}
        \end{minipage}
    }
    \hspace{0.05 \linewidth}
    \subfigure[Zero123++ \cite{shi2023zero123++} fails to generate consistent results in terms of pose and appearance.]{
        \begin{minipage}{0.45\linewidth}
            \centering
            \includegraphics[width=0.9\linewidth]{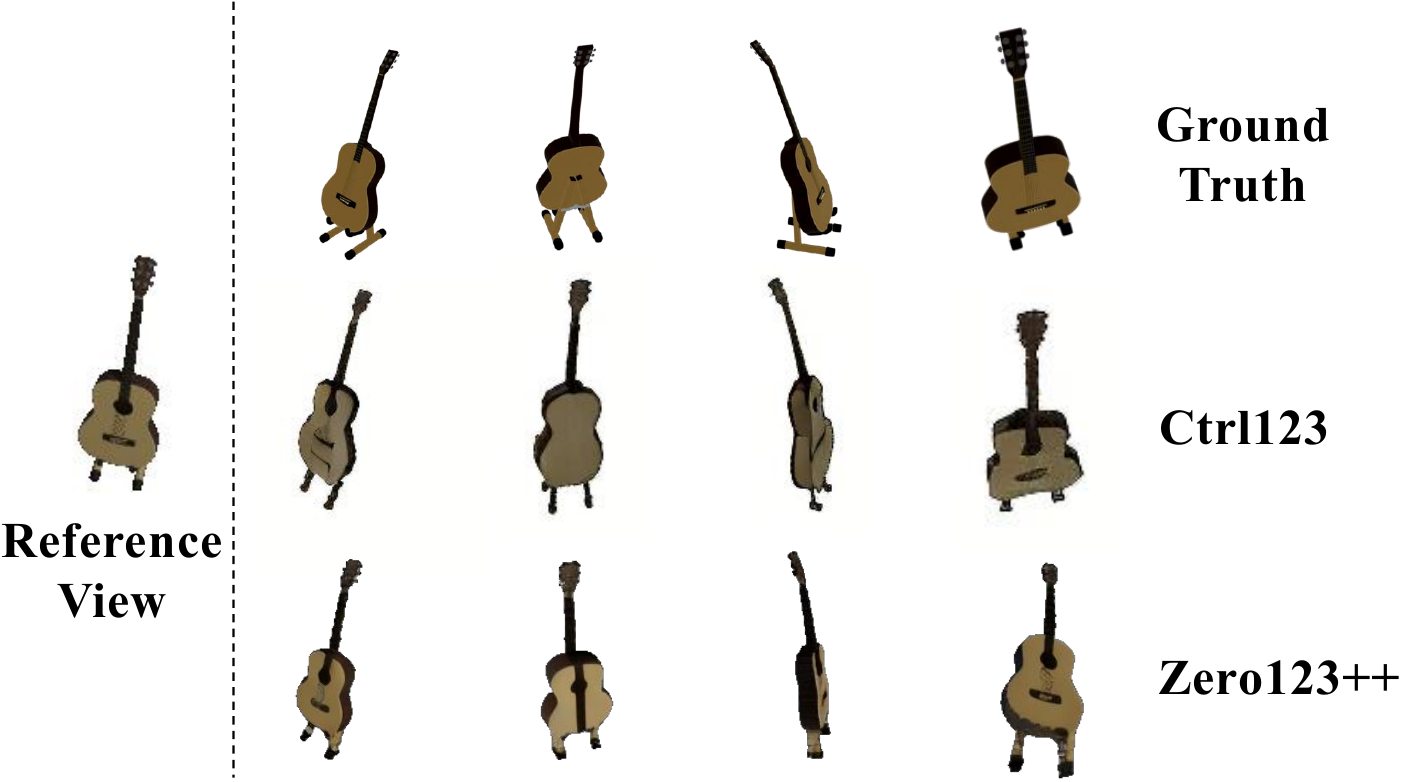}\label{fig:fail-case-b}
        \end{minipage}
    }
    \caption{A qualitative comparison between the generated novel views and their corresponding ground truth for an object from the \textit{training set}. Both Zero123-XL \cite{deitke2023objaverse} and Zero123++ \cite{shi2023zero123++} fail to generate results highly consistent with the ground truth in terms of pose and appearance, while \ours{} can significantly improve consistency in the generated novel views.}\label{fig:fail_case_of_zero123}
    \vspace{-0.7cm}
\end{figure}

Nevertheless, there still exists a fundamental problem in the current NVS models -  it is difficult to ensure the generated views to be  \textit{``consistent''} with the desired poses and appearances (the ground truth), even on the training images, as demonstrated in Figure~\ref{fig:fail_case_of_zero123}. Both Zero123-XL \cite{deitke2023objaverse} and Zero123++ \cite{shi2023zero123++} fail to generate novel views with 1) accurate poses closely aligned with the ground truth and 2) high-quality image details, which indicate that this inconsistency problem cannot be resolved by simply scaling up the training dataset or simplifying the task setting.

The inconsistency problem is rooted in the training strategy of diffusion models which separately trains the denoiser at different noise levels. More specifically, in each training step of diffusion models, a random timestep $t \in [0, T]$ is sampled with a pre-defined noising schedule  $\bar{\alpha}_t \in (0,1)$ to corrupt a clean image $\x$ to a noised image $\x_t$ according to $\mathbf{x}_{t} =\sqrt{\bar{\alpha}_t} \mathbf{x} + \sqrt{1-\bar{\alpha}_{t}}\boldsymbol{\epsilon}$, where $ \boldsymbol{\epsilon} \sim \mathcal{N}(\mathbf{0}, \mathbf{I})$.
Then, a U-Net based conditional denoiser $\phi(\x_t, \e; \omega)$ parameterized by $\omega$, takes $\x_t$ and the condition $\e$ as input and predicts the added noise $\hat{\mathbf{\epsilon}}$. The denoiser is trained with the score-matching loss $\mathop{\min}_{\omega} \| \phi(\x_t, \e; \omega) - \mathbf{\epsilon} \|_2^2$, at each single noise level $t$, without any constraints on the whole denoising process (from Gaussian noise $\x_T$ to the clean image $\x_0$). This strategy significantly improves the diversity of the generated images but neglects the consistency required by NVS models. In this work, we explore ways to enforce consistency between the generated views and ground truth to enhance the capability of NVS models. 

One straightforward way to enforce sample-wise consistency is to add a loss such as Mean Square Error (MSE) in the \textit{pixel space} between the generated views and ground truth. However, through extensive experiments, a direct loss in pixel space often leads to training collapse due to the nature of Diffusion Models (see Appendix~\ref{appendix:ablation} for more details). Inspired by the recently proposed closed-loop transcription (CTRL) framework~\cite{dai2022ctrl, ma2022principles, tong2022incremental}, in this work, we propose to align poses of the generated views and ground truth in a \textit{latent space} using patch features which includes fine-grained information. We name our method \ours{}, a CTRL-based novel view synthesis method that significantly alleviates the problem of pose inconsistency in NVS. Generally speaking, most current single-image NVS methods~\cite{liu2023zero,shi2023zero123++,weng2023consistent123} can be viewed as an open-loop auto-encoder (as shown in Figure \ref{fig:diagram-a}). In this work, we extend the open-loop framework to a closed-loop one by feeding the generated views into the encoder (as shown in Figure \ref{fig:diagram-b}). We then measure the difference between generated views and its corresponding ground truth in the latent feature space. To quantitatively measure the pose consistency of the generated views, we introduce metrics (Average Angle (AA) and Intersection over Union (IoU)) into NVS to measure fine-grained and coarse levels of pose-consistency. Through extensive experiments, we show that \ours{} significantly improves the NVS pose and appearance consistency which further leads to significantly better 3D reconstruction compared to the current SOTA methods. 
The main contributions of this paper are:
\begin{itemize}
    \item To improve pose and appearance consistency in NVS, we introduce \ours{}, a novel closed-loop transcription-based novel view synthesis diffusion model that uses patch features to measure (and minimize) differences between generated views and the ground truth.
    \item An in-depth experiment on a sample set of 25 training objects demonstrates that \ours{} exhibits significant superiority to Zero123 \cite{liu2023zero}, generating novel views much more consistent with the ground truth. Quantitatively, \ours{} improves the quality of generated views by a \textit{7 point} increase in PSNR. Furthermore, \ours{} significantly improves the NVS consistency with  a\textit{ 35.1\% }increase in AA$^{15^{\circ}}$ and \textit{42.5\% }increase in IoU$^{0.7}$ (see Table~\ref{tab:ablation_rounds_number}).
    \item Encouraged by these findings, we further train \ours{} on a large-scale 3D dataset Objaverse~\cite{objaverse} and observe similar improvements. \ours{} improves AA and IoU on two evaluation datasets, with 2.5\%, 4.9\% improvement on AA$^{15^{\circ}}$, and 14.8\%, 9.5\% improvement on IoU$^{0.7}$, respectively.
    
\end{itemize}

\section{Related Works}
\label{sec:related}

\subsection{Diffusion Models}
Diffusion models~\cite{ddpm} demonstrate remarkable capability in image generation. Together with the development of Vision-Language Models such as CLIP~\cite{radford2021learning}, it accelerates text-to-image~\cite{dalle2, imagen, GLIDE, latentdiffusion} models on large-scale web data~\cite{cc12m,laion400m,laion5b}. Subsequent works leverage the powerful pre-trained diffusion models for more controllable image generation, conditioning on more fine-grained prompts such as layout, sketch, and human pose~\cite{gligen, controlnet}. Recent studies have explored distilling knowledge from pre-trained diffusion models for text-to-3D generation through optimizing a differentiable 3D representation with image priors~\cite{poole2022dreamfusion, SJC,chen2023fantasia3d, lin2023magic3d,wang2023prolificdreamer,huang2023dreamtime}. However, this line of work suffers from optimization efficiency, resulting in blurriness and the Janus problem in the generated 3D models. This arises due to the lack of 3D-awareness in pre-trained text-to-image models.
\subsection{Novel View Synthesis with Diffusion Models}
Novel view synthesis (NVS) requires generating an object's unobserved geometry and texture, which is a prerequisite capability for 3D generation. Generative models, such as diffusion models~\cite{latentdiffusion,ddpm} are well-suited for NVS due to their impressive generation capabilities. This is especially advantageous when dealing with sparse input views, to the extreme with a single image. Recently, Zero123~\cite{liu2023zero} proposes to perform zero-shot open-set NVS by fine-tuning a pre-trained text-to-image diffusion model on multi-view renderings of diverse 3D data~\cite{objaverse}. However, this fine-tuning process is computationally expensive and data-demanding. It remains challenging to collect large amounts of high-quality 3D data. As a result, Zero123 \cite{liu2023zero} often generates inconsistent poses compared to the ground truth. Subsequent efforts ~\cite{liu2023syncdreamer,weng2023consistent123,shi2023mvdream,long2023wonder3d} aim to resolve view inconsistency by facilitating information propagation across different views which naturally require generating multiple novel views concurrently. Typically, these methods construct training data by fixing an evenly spaced set of camera poses with fixed elevation, and stacking multiple views together to train attention modules that model multi-view dependencies. While these models excel in generating more multi-view consistent novel views, they face challenges in generalizing to arbitrary camera poses and remain 3D-data demanding. 

\vspace{-0.1cm}
\section{Method}
\label{sec:method}

\begin{figure}
    \vspace{-0.7cm}
    \centering
    \subfigure[Existing NVS models (Zero123 \cite{liu2023zero} and etc.) can be largely viewed as an (open-loop) auto-encoder.]{
        \begin{minipage}{0.5\linewidth}
            \centering
            \includegraphics[width=0.9\linewidth]{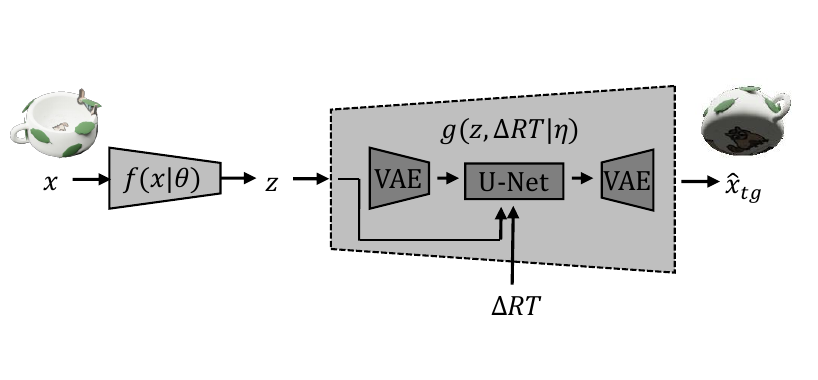}\label{fig:diagram-a}
        \end{minipage}
    }
    \hspace{0.05\linewidth}
    \subfigure[\ours{} extends the open-loop NVS models to a closed-loop  framework (transcription).]{
        \begin{minipage}{0.35\linewidth}
            \centering
            \includegraphics[width=0.9\linewidth]{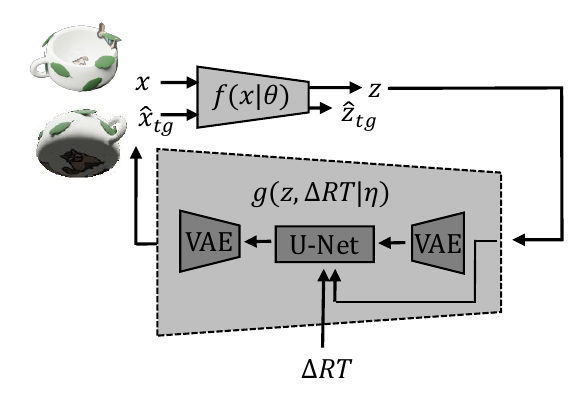}\label{fig:diagram-b}
        \end{minipage}
    }
    \vspace{0.0001\linewidth}
    \caption{Comparison between the training pipeline of current open-loop NVS models and closed-loop \ours{}.}\label{fig:diagram}
    \vspace{-0.5cm}
\end{figure}

Our goal is to learn a novel view synthesis (NVS) model that alleviates the ``inconsistency'' problem by employing the idea of closed-loop transcription~\cite{dai2022ctrl,ma2022principles}. To this end, we formulate the current state-of-the-art (SOTA) methods for single-image NVS \cite{liu2023zero,shi2023zero123++,weng2023consistent123} as an open-loop auto-encoder framework (Section \ref{sec:method:autoencoder} and Figure~\ref{fig:diagram-a}). Then, we extend the open-loop to closed-loop to learn the decoder/generator with the help of the closed-loop transcription (CTRL) framework (Section \ref{sec:method:ctrl123} and Figure~\ref{fig:diagram-b}). Finally, the choice of training strategy and denoise scheduler are discussed in Section \ref{sec:implement}.

\subsection{Current NVS Methods and their Caveats}
\label{sec:method:autoencoder}
\paragraph{The Formulation of Current NVS Methods.}
Given a single RGB image (reference view) of an object, the goal of single image NVS is to synthesize an image of the object from a different camera viewpoint (target view) given the relative camera transformation. Although various works improve the NVS task in different aspects, we could unify their formulation as an open-loop auto-encoder as follows.

Suppose the random variables $\X$, $\Delta \bm{RT}$, and $\X_{tg}$ denote the reference image, the relative camera extrinsic, and the target image, respectively. The dataset $\mathcal{D}$ comprising $n$ triplets represented as $\mathcal{D}= \left\{ \left(\X^i, \Delta \bm{RT}^i, \X^i_{tg}\right)\right\}_{i=1}^n$, where the given $n$ i.i.d. samples $\X^1,\dots, \X^n \sim \X$, $\Delta \bm{RT}^1,\dots, \Delta \bm{RT}^n \sim \Delta \bm{RT}$, and  $\X_{tg}^1,\dots, \X_{tg}^n \sim \X_{tg}$. Specifically, $\Delta \bm{RT}^i = (\Delta \bm{R}^i, \Delta \bm{T}^i)$, where $\Delta \bm{R}^i \in \Re^{4\times 3}$ and $\Delta \bm{T}^i \in \Re^{4}$ respectively represent the relative camera rotation and translation. The current single image NVS methods aim to learn a model $h$ that synthesizes new image $\hat{\X}_{tg}$ from the reference image $\X$ under the relative camera transformation $\Delta \bm{RT}$, i.e,
\begin{align}
\hat{\X}_{tg} = h(\X, \Delta \bm{RT}),
\end{align}
where $\hat{\X}_{tg}$ is the synthesized target view image. The goal is for $\hat{\X}_{tg}$ to be perceptually similar to the true target view $\X_{tg}$.


The model $h$ in these methods \cite{liu2023zero,shi2023zero123++,liu2023syncdreamer} is composited by an encoder $f: \X \mapsto \bm{Z}$, parameterized by $\theta$, and a decoder $g: (\bm{Z}, \Delta \bm{RT}) \mapsto \hat{\X}_{tg}$, parameterized by $\eta$, 
i.e., $h(\X, \Delta \bm{RT}) = g(f(\X), \Delta \bm{RT})$. Specifically, the encoder $f$ is commonly chosen to be the ViT \cite{dosovitskiy2020image} model of a pre-trained CLIP \cite{radford2021learning} and kept frozen during the NVS training. Let $\bm{Z} = f(\X;\theta) \in \Re^{d}$, where $\bm{Z}$ is the class feature (cls token) of ViT.

The decoder $g$ is commonly implemented as a process of multi-step conditional latent denoising \cite{liu2023zero} with VAE encoder/decoder (as shown in the dashed box of Figure \ref{fig:diagram-a}). First, the process is conditioned on a transformation $\e = \psi(\bm{Z}, \Delta \bm{RT})$ of the feature $\bm{Z}$ and relative camera extrinsic $\Delta \bm{RT}$, where $\psi$ is a linear layer. Then, for step $t\in \{0,\dots,T\}$, the noise $\hat{\mathbf{\epsilon}}$ is predicted, using an U-net $\phi$, based on the predicted target view image at time step $t$, and the condition $\e$, i.e., $\hat{\mathbf{\epsilon}} = \phi(\hat{\x}_{tg,t}, \e, t)$.  
$\hat{\x}_{tg,t}$ is the latent target view feature $\x_{tg}$ plus $t$ steps noise where $\x_{tg}$ is latent feature of target view image $\X_{tg}$ by VAE encoder.

The multi-step denoising is performed through a denoiser  $\cS(\cdot)$, based on denoising diffusion implicit models (DDIM)~\cite{song2020denoising}. The predicted target view image at time step $t-1$ is generated by the denoiser, i.e., $\hat{\x}_{tg,t-1} = \cS(\hat{\x}_{tg,t}, \hat{\mathbf{\epsilon}}, t)$ \footnote{$\cS(\hat{\x}_{tg,t}, \hat{\mathbf{\epsilon}}, t)=\frac{1}{\sqrt{\bar{\alpha}_t}} (\hat{\x}_{tg,t} - \frac{1-\alpha_t}{\sqrt{1-\bar{\alpha}_t}} \hat{\mathbf{\epsilon}})$ where $\bar{\alpha}_t = \prod_{i=1}^{t}\alpha_i$ and $\alpha_t=1-\beta_t$, $\beta_t$ is a pre-defined variance of $t$-th step.}. For the initial step $t=t_\infty$, the $\hat{\x}_{tg,t_\infty}$ is randomly sampled from isotropic Gaussian. Then, after $t_\infty$ denoising steps, we get $\hat{\x}_{tg,0}=\hat{\x}_{tg}$. Finally, the denoised latent feature $\hat{\x}_{tg}$ was lifted to generate target view image $\hat{\X}_{tg}$ through the VAE decoder.

We denote random variable $\hat{\X}_{tg} \doteq g(\bm{Z}, \Delta \bm{RT})$ as the decoded from Gaussian noise $\hat{\x}_{tg,t_\infty}$ through $t_\infty$ denoising steps according to $\cS$ conditioned on $\bm Z$ and $\bm{\Delta RT}$. Current single-image NVS methods can be summarized as the following open-loop framework,
\begin{equation}
    \X \xrightarrow{f(\X; \theta)\hspace{2mm}} \bm Z \xrightarrow{\hspace{2mm} g(\bm{Z}, \Delta \bm{RT};\eta) } \hat{\X}_{tg}.
    \label{eqn:zero123}
\end{equation}
Such diffusion-based NVS methods typically adopt a standard diffusion training strategy by supervising single-step denoising results with a score-matching loss, i.e.,  
\begin{equation}
    \mathop{\min}_{\eta} \mathbb{E}\| \phi(\hat{\x}_{tg,t}, \psi(\bm{Z}, \Delta \bm{RT}), t) - \mathbf{\epsilon} \|_2^2,
    \label{eqn:sm}
\end{equation}
where the expectation is taken over the encoded feature $\bm{Z}$, timestep $t$, relative camera extrinsic $\Delta \bm{RT}$, and randomly sampled Gaussian noise $\mathbf{\epsilon}$.

\paragraph{The Caveats of Current NVS Methods.}
While the showcased results in most single-image NVS works appear impressive, we make a critical observation previously overlooked: the generated novel views often lack consistency with the ground truth, even when evaluated on the training data, as exemplified in Figure~\ref{fig:fail_case_of_zero123}. This inconsistency problem is rooted in the training strategy of diffusion models which separately trains the denoiser at different noise levels using solely score matching loss without any constraints on the whole denoising process (from Gaussian noise $\hat{\x}_{tg,t_\infty}$ to the clean image $\hat{\X}_{tg}$). This strategy significantly improves the diversity of the generated images but neglects the consistency required in NVS models. In this work, we explore ways to enforce consistency between the generated views and ground truth to enhance the capability of NVS models.

\subsection{\ours{}: Consistency via a Closed-Loop Framework}
\label{sec:method:ctrl123}

A straightforward method to enforce sample-wise consistency is to add a loss such as Mean Square Error (MSE) in the \textit{pixel space} between the generated view $\hat{\X}_{tg}$ and ground truth $\X_{tg}$. However, through extensive experiments, a direct loss in pixel space often suffers training difficulties leading to training collapse (See Appendix~\ref{appendix:ablation} and Table 
\ref{tab:ablation_consistency_type} for more details).

The recently proposed {\em closed-loop transcription} (CTRL) framework~\cite{dai2022ctrl} is naturally suited to solve this problem. The difference between $\X$ and $\hat \X$ can be measured through the distance between their corresponding features $\bm Z = f(\bm{X}; \theta)$ and $\hat{\bm Z} = f(\hat{\bm{X}}; \theta)$ mapped through the same encoder, i.e., 
\begin{equation}
\X \xrightarrow{\hspace{2mm} f(\X; \theta)\hspace{2mm}}  \bm Z \xrightarrow{\hspace{2mm} g(\bm{Z};\eta) \hspace{2mm}} \hat \X \xrightarrow{\hspace{2mm} f(\hat \X; \theta)\hspace{2mm}} \ \hat {\bm Z}.
\label{eqn:CTRL}
\end{equation}

Inspired by CTRL, we apply this idea to the current NVS model that measures the difference between $\X_{tg}$ and the generated $\hat{\X}_{tg}$ in the CLIP feature space. In other words, we apply it to framework \ref{eqn:zero123} to get the following new framework:
\begin{equation}
\X \xrightarrow{\hspace{2mm} f(\X; \theta)\hspace{2mm}}  \bm Z \xrightarrow{\hspace{2mm} g(\bm{Z},\Delta \bm{RT};\eta) \hspace{2mm}} \hat{\X}_{tg} \xrightarrow{\hspace{2mm} f(\hat{\X}_{tg}; \theta)\hspace{2mm}} \ \hat{\bm Z}_{tg}.
\end{equation}

Different from the direct CTRL formulation \eqref{eqn:CTRL}, we can not directly calculate the loss between $\bm Z$ and $\hat{\bm Z}_{tg}$ since they are the features of different views. Hence, we add the feature $\bm Z_{tg}$ of the ground truth target view $\X_{tg}$  as the following,
\begin{equation}
\begin{aligned}
    \X \xrightarrow{f(\X; \theta)\hspace{2mm}} \bm Z\xrightarrow{\hspace{2mm} g(\bm{Z}, \Delta \bm{RT};\eta) } \hat{\X}_{tg} &\xrightarrow{f(\hat{\X}_{tg}; \theta)\hspace{2mm}} \hat{\bm Z}_{tg}, \\
    \X_{tg} &\xrightarrow{f(\X_{tg}; \theta)\hspace{2mm}} {\bm Z}_{tg}. 
    \label{eqn:ctrl123_1}
\end{aligned}
\end{equation}

Now we can measure the $\bm Z_{tg}$ and $\hat{\bm Z}_{tg}$ through Mean Squared Error, 
\begin{equation}
\mathop{\min}_{\eta} \| \bm Z_{tg} - \hat{\bm Z}_{tg} \|_2^2.
\end{equation}

Building upon the previously established definition of $\bm{Z}$, we identify it as the class feature within the outputs of the Vision Transformer (ViT) model. Revisiting the ViT model, its outputs are bifurcated into class features and patch features. The class features capture high-level information, whereas the patch features are indicative of low-level information. Our methodology prioritizes the analysis of patch features due to their richer and more detailed informational content, which significantly aligns with ground truth data and enhances consistency. This preference is empirically validated in our ablation study, which demonstrates superior performance of patch features over class features (refer to Appendix~\ref{appendix:ablation} and Table \ref{tab:ablation_feature_type} for detailed results).

To accommodate this distinction, we introduce a revised notation for the encoder outputs, denoted as $\left[\bm{Z}_c, \bm{Z}_p\right] = f(\mathbf{X}, \theta)$ and $\left[\bm{Z}_{tg,c}, \bm{Z}_{tg,p}\right] = f(\mathbf{X}_{tg}, \theta)$, where $\bm{Z}_c$ and $\bm{Z}_{tg,c}$ represent the high-level class features, and $\bm{Z}_p$ and $\bm{Z}_{tg,p}$ correspond to the low-level patch features. Re-writing the framework \ref{eqn:ctrl123_1} with the re-defined notation as the following,
\begin{equation}
\begin{aligned}
    \X \xrightarrow{f(\X; \theta)\hspace{1mm}} \bm Z_c \xrightarrow{\hspace{1mm} g(\bm{Z}_c, \Delta \bm{RT};\eta) } \hat{\X}_{tg} &\xrightarrow{f(\hat{\X}_{tg}; \theta)\hspace{2mm}} \hat{\bm Z}_{tg,p} \\
    \X_{tg} &\xrightarrow{f(\X_{tg}; \theta)\hspace{2mm}} {\bm Z}_{tg,p}, 
    \label{eqn:ctrl123_2}
\end{aligned}
\end{equation}
where $\bm Z_c$ is the class feature of reference views $\X$ and ${\bm Z}_{tg,p}$, $\hat{\bm Z}_{tg,p}$ are the patch features of target view ground truth $\X_{tg}$ and generated target view $\hat{\X}_{tg}$, and we can measure the difference between $\bm Z_{tg,p}$ and $\hat{\bm Z}_{tg,p}$ through Mean Squared Error, 
\begin{equation}
\mathop{\min}_{\eta} \| \bm Z_{tg,p} - \hat{\bm Z}_{tg,p} \|_2^2.
\label{eqn:closedloop_loss}
\end{equation}

\paragraph{Alternative Training Strategy}
\label{sec:implement}


Instead of directly applying \ref{eqn:closedloop_loss} as a regularization term over the score-matching loss, we propose an alternative training strategy which is ablated in Table \ref{tab:ablation_opt_strategy}. We perform fine-tuning on pre-trained Zero123 \cite{liu2023zero} (denoted as Zero123 in the rest part of our paper).
One round of training involves 1) $m$ iterations of closed-loop training via CTRL framework as depicted in \ref{eqn:ctrl123_2} with supervision loss according to \ref{eqn:closedloop_loss}. We call this process CL training; 2) $n$ iterations of standard diffusion model fine-tuning with score-matching loss. We call this process SM training for short.  Our experiment results show that more rounds of such training could continuously improve NVS performance. Such an alternative training strategy is more efficient and practical as the hyper-parameters of CL training and SM training are vastly different. It is time-consuming and energy-inefficient to grid search a sweet spot of hyper-parameters for both CL and SM training. In section \ref{sec:exp:ablation} we demonstrate the effectiveness of this alternative training strategy.

\vspace{-0.1cm}
\section{Experiments}

We evaluate \ours{} on the tasks of single-view NVS and 3D reconstruction. 3D reconstruction is a more challenging task that requires strong multiview consistency. In Section \ref{sec:exp:settings}, we describe our data collating and the implementation details. We introduce the metrics used to measure pose consistency between the generated novel views and the ground truth in Section~\ref{sec: AA}, including our introduced metric Angle Accuracy (AA) and Intersection over Union (IoU). We provide in Section~\ref{sec:exp:largescale} results on both small-scale and large-scale experiments, trained on a subset of the curated dataset and the full dataset, respectively. Results on 3D reconstruction are provided in Section~\ref{sec:exp:3D}.  Ablations on the design choices are presented in Section~\ref{sec:exp:ablation}.

\begin{table}[htb]
\vspace{-0.2cm}
\centering
    \small
    \caption{Ablation study on the number of rounds for alternative training strategy when trained on a subset of the curated dataset, measured with 4 metrics: KID($\downarrow$), PSNR($\uparrow$), AA$^{x^{\circ}}$($\uparrow$), and IoU$^{x}$($\uparrow$). While Zero123-small shows an unnoticed problem when trained on the subset dataset - Zero123-small generates pose-inconsistent novel views, Ctrl123-small produces significantly more consistent results which continue to improve with more rounds of alternative training. Please refer to Appendix~\ref{appendix:num-rounds-viz}  for more visualization results.}
    \label{tab:ablation_rounds_number}
    \begin{tabular*}{\hsize}{@{}@{\extracolsep{\fill}}llllllll@{}}
    \toprule
    Method vs.  & Training & \multirow{2}{*}{Image Quality} &  Alignment  \\ 
    Metrics     & steps  &  & (MegaPose/DIS)  \\ 
    \midrule

    \multirow{2}{*}{Zero123-small} & \multirow{2}{*}{20,000} & KID:0.0254 & AA$^{15^{\circ}}$:22.62\% \\
                            &  & PSNR:19.3839 & IoU$^{0.7}$ :30.93\% &  \\
    \midrule
    \ours{}-small & \multirow{2}{*}{10,000} & KID:0.0087 & AA$^{15^{\circ}}$:32.02\% \plusvalue{9.40\%}  \\
    (1 round)   & & PSNR:23.6080 &  IoU$^{0.7}$ :55.01\% \plusvalue{24.08\%}  \\
    \midrule
    \ours{}-small & \multirow{2}{*}{20,000} & KID:0.0067 & AA$^{15^{\circ}}$:57.78\% \plusvalue{35.16\%}   \\
    (2 rounds)   & & PSNR:26.5348 & IoU$^{0.7}$ :73.44\%  \plusvalue{42.51\%}   \\
    \bottomrule
    \end{tabular*}%
    \vspace{-0.3cm}
\end{table}


\subsection{Settings}
\label{sec:exp:settings}

\paragraph{Dataset and Baseline.} We use Zero123 as our baseline, and utilize the public 3D dataset - Objaverse \cite{objaverse} which contains around 800K diverse 3D models created by artists. For training efficiency and quality, we apply the data curation similar to  \cite{li2023instant3d}, eliminating 3D assets with sub-optimal geometry, low-quality texture, or small sizes. This curation process results in a refined training dataset with approximately 100K 3D models. 
All other configurations follow the settings in the literature of baseline (e.g. the number of view sampling, sampling strategy, and camera model, among others).

\paragraph{Implementation.} We train \ours{} on 100K curated Objaverse 3D models for 2 rounds of alternative training. We initialize \ours{} with pre-trained weights of Zero123 and train on 8 A100 GPUs with 80GB memory. For each alternative training round, we conduct 500 steps of closed-loop (CL) training with a total batch size of 320 and a learning rate of $10^{-5}$, followed by $t_\infty=1000$ steps of score-matching (SM) training with a total batch size of 1536 and a learning rate of $10^{- 4}$, each taking around 2 days. We use gradient accumulation to increase the training batch size, 160 for CL training and 2 for SM training respectively. During model training, we use the Adam optimizer \cite{Kingma2014AdamAM} with $\beta_1=0.9$ and $\beta_2=0.99$. In all the experiments, we train our model with the 16-bit floating point (fp16) format for efficiency. 

\begin{table*}[t]
\vspace{-0.7cm}
\centering
    \scriptsize
    \setlength{\tabcolsep}{3pt}
    \caption{Quantitative comparison of \ours{} and four baselines, measured with 4 different metrics: PSNR ($\uparrow$), KID ($\downarrow$), AA$^{x^{\circ}}$($\uparrow$), and IoU$^{x}$($\uparrow$), 
     and evaluated on 3 datasets: GSO, RTMV, and 25 randomly sampled objects from the training set.}
    \label{tab:compare_with_zero123}
    \begin{tabular*}{\hsize}{@{}@{\extracolsep{\fill}}lcccccccccccc@{}}
    \toprule
    \multirow{2}{*}{Method} & \multicolumn{4}{c}{GSO} & \multicolumn{4}{c}{RTMV} & \multicolumn{4}{c}{25 Training Objects} \\
    \cline{2-5} \cline{6-9} \cline{10-13}     
    & PSNR & KID & AA$^{15^{\circ}}$ &IoU$^{0.7}$
    & PSNR & KID & AA$^{15^{\circ}}$ &IoU$^{0.7}$ & PSNR & KID & AA$^{15^{\circ}}$ &IoU$^{0.7}$\\
    \midrule
    Zero123 & 17.2049 & 0.0055 & 14.41\% & 47.32\%&  8.9123&  0.0340& 7.04\% & 2.91\%& 15.9452 & \textbf{0.0043} & 14.28\% & 31.86\%\\
    SyncDreamer \cite{liu2023syncdreamer} & 17.6317 & 0.0077 & 15.94\% & 51.65\% & / & / & / & / & / & / & / & /\\
    Zero123++ \cite{shi2023zero123++} & 18.2247 & 0.0140 & 15.63\% & 32.45\% & / & / & / & / & / & / & / & /\\
    Consistent123 \cite{weng2023consistent123} & 13.8337 & 0.0456 & 3.17\% & 29.85\% & / & / & / & / & / & / & / & /\\
    \ours{} & \textbf{18.4712} & \textbf{0.0045} & \textbf{16.96\%} & \textbf{62.11\%}& \textbf{10.8938} & \textbf{0.0294} &\textbf{11.98\%} & \textbf{12.41\%}& \textbf{18.9398} & 0.0052 & \textbf{19.79\%} &\textbf{45.97\%} \\
     \bottomrule
    \end{tabular*}%
    \vspace{-0.3cm}
\end{table*}

\begin{figure*}[t]
  \centering
   \includegraphics[width=0.75\linewidth]{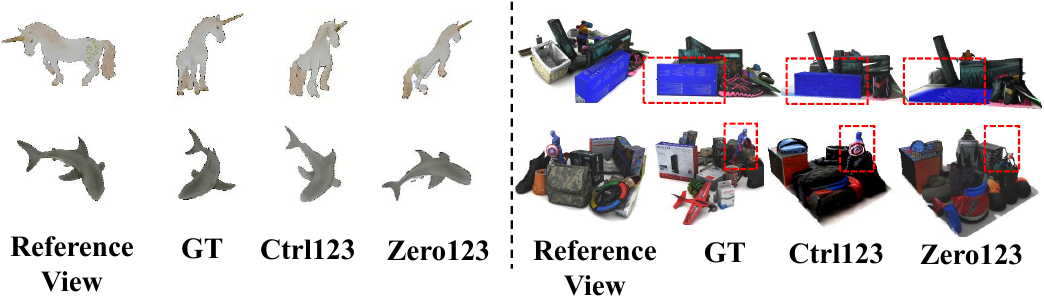}
   \caption{Qualitative comparison of NVS generalization capability on GSO (Left) and RTMV (Right) after training on large-scale dataset (100K). More cases can be found in Appendix~\ref{appendix:train_on_100k}.}
   \label{fig:ctrl2zero123}
   \vspace{-0.7cm}
\end{figure*}


\subsection{Angle Accuracy (AA) and Intersection over Union (IoU)} \label{sec: AA}
The uniqueness of NVS makes the evaluation a multifaceted problem. We cannot \textit{only} evaluate the model's performance via traditional image generation benchmarks such as KID, FID, PSNR, and SSIM. As discussed in Sections \ref{sec:intro} and \ref{sec:method}, it is equally important to evaluate the model's ability to generate pose ``consistent" views. In doing so, we introduce two metrics \textit{Angle Accuracy (AA)} and \textit{Intersection over Union (IoU)} that measure pose consistency on both \textit{fine-grained} and \textit{coarse} level.

\vspace{-0.2cm}
\paragraph{Angle Accuracy (AA).} AA is first introduced in \cite{zhou2019neurvps}. For each camera pose predicted in the generated novel view, we calculate the angle difference between the ground truth and the predicted pose. The development of the 6D pose estimation task enables the prediction of 6D pose from a single image with its GT CAD model. In this work, we use MegaPose \cite{labbe2023megapose} which is a single image pose estimator, to estimate the camera pose of the generated view. We then identify the percentage of predictions as AA$^{x^{\circ}}$ where the angle difference falls within a predefined threshold $x^{\circ}$. To varify the credibility of AA, we calculate multiple AA$^{x^{\circ}}$ when $x$ is set as 5,10,15,20, which all reflects better pose alignment of Ctrl123. Please refer to Appendix \ref{sec: AA credibility} to see more illustrations for the credibility of our introduced AA. After careful comparison and analysis, we choose AA$^{15^{\circ}}$ as the final metric to report in the comparison tables. The numerical analysis for the chosen AA$^{15^{\circ}}$ can be found in Appendix \ref{sec: AA credibility}.


\vspace{-0.2cm}
\paragraph{Intersection over Union (IoU).} We also employ a segmentation metric IoU \cite{garcia2017review} that captures more coarse aspect of pose consistency. We use DIS \cite{qin2022highly} to predict the mask of the generated view and the ground truth view to measure the IoU. The calculation of IoU is defined by dividing the overlap between the predicted and ground truth mask by the union of them.

For all experiments in this paper, we report KID($\downarrow$), PSNR($\uparrow$), AA$^{x^{\circ}}$($\uparrow$) and IoU$^{x}$($\uparrow$) to compare the high-level quality, low-level details, and pose consistency of the generated novel views among all methods. ($\uparrow$) denotes the higher the better, verse vice for ($\downarrow$)). The PSNR metric is an overall metric of texture and silhouette and it is not linearly related to silhouette. We also introduce IoU metric which evaluates silhouette directly and reflects coarse pose alignment.

\begin{figure}
    \centering
    \begin{minipage}[b]{0.45\linewidth}
        \centering
        \includegraphics[width=1.0\linewidth]{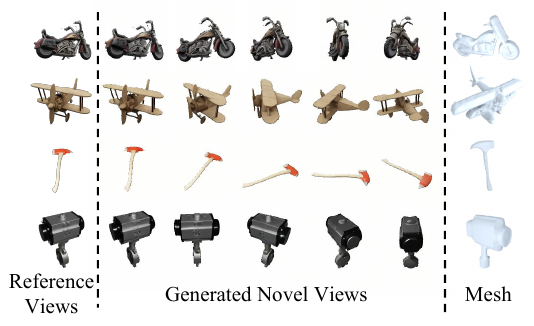}
        \caption{NVS/3D examples using \ours{} on images from training set (100K).}
        \label{fig:3d-train}
    \end{minipage}
    \hspace{0.05\linewidth}
    \begin{minipage}[b]{0.45\linewidth}
        \centering
        \includegraphics[width=0.9\linewidth]{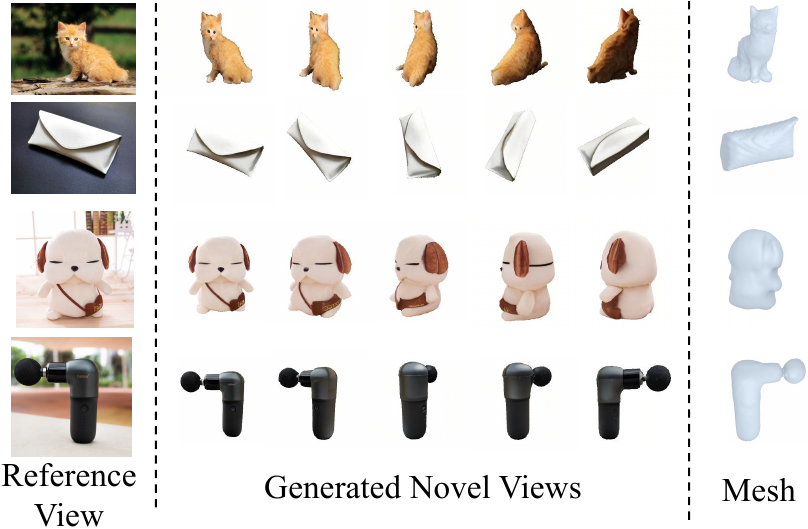}
        \caption{NVS/3D examples using \ours{} on images in-the-wild.}
        \label{fig:3d-test}
    \end{minipage}
    \vspace{-0.7cm}
\end{figure}

\subsection{Comparison with SOTA}
\label{sec:exp:largescale}

\paragraph{Improvement with Multi-rounds Optimization.} First, we evaluate the performance of multi-rounds training strategy of \ours{} on a small dataset. The dataset consists of 25 objects from Objaverse, each with 12 randomly sampled views. We first train a baseline (Zero123-small in Table~\ref{tab:ablation_rounds_number}), which is initialized with pre-trained weights of Stable Diffusion, and then fine-tuned on the 25 objects by score matching loss for 20,000 iterations with batch size of 96. We train \ours{}-small following the alternative training strategy as detailed section~\ref{sec:implement}. For fair comparison, it is initialized with the checkpoint of Zero123-small at 10,000 iterations. We train for 500 closed-loop iterations and 4,500 score-matching iterations as one round. Therefore, after two rounds of training, \ours{}-small will accumulate the same number of training iterations as the baseline (20,000).

An interesting problem happens: even if Zero123-small has already converged, its pose alignment is worse than that of Zero123 (based on the comparison of metrics: AA$^{x^{\circ}}$ and IoU$^{x}$), as shown in Appendix \ref{appendix:ablation} and Table \ref{tab:zero123 vs small}. Therefore, The training strategy of Zero123 will cause pose inconsistency, which is more obvious when training on a small dataset. As shown in Table~\ref{tab:ablation_rounds_number}, \ours{}-small solves this problem and significantly improves over Zero123-small both qualitatively and quantitatively (more quantitative results in Appendix \ref{appendix:ablation} and Figure \ref{fig:C1}). After 2 rounds of training, \ours{}-small achieves better pose-alignment.
The PSNR increases by 7.2 points which is almost 2$\times$ better than baseline \footnote{Based on the definition of PSNR, 3 points equals 1$\times$ visual quality improvement \cite{wang2004image}}. \ours{} also generates more pose-consistent views. On the fine-grained level, AA increases from 22.62\% to 57.78\%; One a more coarse level, IoU increases from 30.94\% to 73.44\%.

\vspace{-0.2cm}
\paragraph{Improvements extend to larger datasets.} Following Zero123 \cite{liu2023zero}, we evaluate on GSO \cite{downs2022google} and RTMV \cite{tremblay2022rtmv} for NVS quality in Table~\ref{tab:compare_with_zero123}. We evaluate 20/110 randomly selected objects/scenes for 17/50 sampled views from GSO/RTMV respectively.
We compare with Zero123 on GSO and RTMV , and compare with SyncDreamer \cite{liu2023syncdreamer}, Zero123++ \cite{shi2023zero123++} and Consistent123 \cite{weng2023consistent123} on GSO.\footnote{We also evaluate their (Zero123++, SyncDreamer, and Consistent123) performance on RTMV \cite{tremblay2022rtmv} under similar settings and put the results in Appendix \ref{appendix:train_on_100k} and Table \ref{tab:compare_with_zero123_appendix}. Note that RTMV \cite{tremblay2022rtmv} only offers images of randomly sampled views, which is not fair for these work when evaluating.} Note that we only compare with works under similar settings. Therefore, Wonder3D \cite{long2023wonder3d}, which takes RGB images and normal images as training data, is not in our comparing list.
As shown in Table~\ref{tab:compare_with_zero123}, \ours{} outperforms baselines in all four metrics (PSNR, KID, AA$^{x^{\circ}}$, and IoU$^{x}$) on evaluation datasets. In addition, \ours{} fits better on the subset of training dataset (randomly selected 25 training objects). In particular, \ours{} obtains much better pose-consistency (see AA$^{x^{\circ}}$, and IoU$^{x}$). We visualize the differences in Figure~\ref{fig:ctrl2zero123} for qualitative comparison between \ours{} and Zero123. 
\ours{} generates higher quality and more pose-consistent views comparing to the baseline. 

Note that the code base we use to train and inference is threestudio \cite{threestudio2023}, which select a more efficient but lower quality inference method. Therefore, metrics of Zero123 reported in Table \ref{tab:compare_with_zero123} are lower than that reported in the literature of Zero123 (see more detailed comparison of different code base in appendix \ref{appendix:ablation} and Table \ref{tab:code base}). The conclusion that \ours{} shows better performance than Zero123 is credible because we evaluate \ours{} and Zero123 with the same code base. In addition, we reproduce the results of SyncDreamer \cite{liu2023syncdreamer}, Zero123++ \cite{shi2023zero123++} and Consistent123 \cite{weng2023consistent123} on the same setting as ours and report them in Table \ref{tab:compare_with_zero123}. 


\subsection{3D Reconstruction}
\label{sec:exp:3D}

We also evaluate 3D reconstruction of \ours{} on Objaverse instances, as well as randomly selected web images. We use the SDS \cite{poole2022dreamfusion} loss implemented by threestudio \cite{threestudio2023} for 3D reconstruction. We then evaluate qualitatively on the exported mesh from reconstructed 3D (refer to Appendix~\ref{appendix:image23d_imp_details} for implementation details and Figure~\ref{fig:comparison_on_3D_nvs} shows the comparison of the generated 3D mesh). \ours{} exhibits better reconstruction quality with smooth surfaces and detailed geometry.

Figure \ref{fig:3d-train} and \ref{fig:3d-test} show the NVS and 3D reconstruction results on both the training set and images in the wild. \ours{} performs higher-fidelity and finer detail 3D reconstruction, and multi-view consistent pose consistent novel view synthesis. Figure~\ref{fig:comparison_on_3D_nvs} shows the NVS/3D mesh comparison with other NVS methods. In comparison, \ours{} achieves the best reconstruction quality with smooth surfaces and detailed geometry.

\subsection{Ablation Study}
\label{sec:exp:ablation}

We conduct ablation studies to verify the importance of multi-step distillation and alternative training strategies (Section~\ref{sec:implement}). We sample a subset of Objaverse, 25 objects with 12 randomly sampled views for the ablation studies.

\vspace{-0.2cm}
\paragraph{Different Training Strategy.} In Section~\ref{sec:implement}, the ``Simultaneous'' implementation of the training strategy involves adding Equation \ref{eqn:closedloop_loss} as a regularization term to the score-matching loss of Zero123, with the aim of optimizing both simultaneously. The results in Appendix \ref{appendix:ablation} and Table~\ref{tab:ablation_opt_strategy} shows the ``Alternative'' strategy is better.

\vspace{-0.2cm}
\paragraph{Denoise Scheduler.} 
 The number of denoising steps $n$ used in the function $g$ to generate $\hat{\X}_{tg}$ is crucial for balancing the quality of $\hat{\X}_{tg}$ against memory consumption. To investigate this, we conducted experiments with different numbers of denoising steps (1, 10, 30, 50) to evaluate their impact on image quality and pose alignment (see more details in Appendix \ref{appendix:ablation} and Table \ref{tab:ablation_denoise_step}). Ultimately, we select 50 denoising steps during CL training.

\vspace{-0.3cm}
 \paragraph{Different Random Seeds.} For the initial step $t=t_\infty$, the $\hat{\x}_{tg,t_\infty}$ is randomly sampled from isotropic Gaussian. Therefore, different seeds lead to different $\hat{\x}_{tg,t_\infty}$. We explore whether different random seeds influence the good pose-consistency performance of Ctrl123, as shown in Appendix \ref{appendix:ablation} and Table \ref{tab:seed}. Results show that the influence caused by different random seeds is negligible.

\section{Conclusion}

In this paper, we introduce \ours{}, a closed-loop transcription-based NVS method that significantly alleviates the problem of pose and appearance inconsistency between the generated view and ground truth in NVS. To quantitatively measure the performance of consistency improved by \ours{}, we introduce metrics AA and IoU. Through extensive experiments, we show that the closed-loop \ours{} significantly improves pose-consistency (and appearance consistency) for NVS, and leads to significantly better 3D reconstruction compared to the current SOTA methods. Note that in \cite{ma2022principles} the closed-loop framework is proposed as a general framework for ensuring consistency. The work of  \cite{dai2022ctrl} has shown that the closed-loop formulation is also effective in ensuring consistency for image classes, besides poses. Hence we believe such a closed-loop framework is necessary to ensure consistency in content generation for other attributes, such as  relative pose between objects, as well as their shapes and textures etc. We will leave such generalization to future investigation.

\section{Broader Impacts} 
\label{sec: impact}

Our work applies a general closed-loop framework to generative models for producing 2D images and 3D content with precise desired attributes. However, our work may raise ethical concerns regarding the potential generation of misinformation. It is important to note that we do not introduce new real-world applications, so the potential harms appear to be minimal.


\section{Limitations}
\label{sec: limit}
We find an unnoticed problem of current NVS methods and our work solved this problem.  Our method requires more computational resources. Nevertheless, our work is an important step to introduce closed-loop framework into diffusion models to enhance generality abilities.


\clearpage
\printbibliography

@article{tong2022incremental,
  title={Incremental learning of structured memory via closed-loop transcription},
  author={Tong, Shengbang and Dai, Xili and Wu, Ziyang and Li, Mingyang and Yi, Brent and Ma, Yi},
  journal={arXiv preprint arXiv:2202.05411},
  year={2022}
}

@article{dai2022ctrl,
  title={CTRL: Closed-Loop Transcription to an LDR via Minimaxing Rate Reduction},
  author={Dai, Xili and Tong, Shengbang and Li, Mingyang and Wu, Ziyang and Psenka, Michael and Chan, Kwan Ho Ryan and Zhai, Pengyuan and Yu, Yaodong and Yuan, Xiaojun and Shum, Heung-Yeung and others},
  journal={Entropy},
  volume={24},
  number={4},
  pages={456},
  year={2022},
  publisher={MDPI}
}

@article{song2023consistency,
  title={Consistency models},
  author={Song, Yang and Dhariwal, Prafulla and Chen, Mark and Sutskever, Ilya},
  journal={arXiv preprint arXiv:2303.01469},
  year={2023}
}

@article{ddpm,
  title={{Denoising Diffusion Probabilistic Models}},
  author={Ho, Jonathan and Jain, Ajay and Abbeel, Pieter},
  journal={{Advances in Neural Information Processing Systems}},
  volume={33},
  pages={6840--6851},
  year={2020}
}

@article{dalle2,
  title={{Hierarchical Text-Conditional Image Generation with CLIP Latents}},
  author={Ramesh, Aditya and Dhariwal, Prafulla and Nichol, Alex and Chu, Casey and Chen, Mark},
  journal={arXiv preprint arXiv:2204.06125},
  year={2022}
}

@article{imagen,
  title={{Photorealistic Text-to-Image Diffusion Models with Deep Language Understanding}},
  author={Saharia, Chitwan and Chan, William and Saxena, Saurabh and Li, Lala and Whang, Jay and Denton, Emily and Ghasemipour, Seyed Kamyar Seyed and Ayan, Burcu Karagol and Mahdavi, S Sara and Lopes, Rapha Gontijo and others},
  journal={arXiv preprint arXiv:2205.11487},
  year={2022}
}

@inproceedings{GLIDE,
  title={{GLIDE: Towards Photorealistic Image Generation and Editing with Text-Guided Diffusion Models}},
  author={Nichol, Alex and Dhariwal, Prafulla and Ramesh, Aditya and Shyam, Pranav and Mishkin, Pamela and McGrew, Bob and Sutskever, Ilya and Chen, Mark},
  booktitle={International Conference on Machine Learning},
  year={2021},
}

@inproceedings{latentdiffusion,
  title={{High-Resolution Image Synthesis with Latent Diffusion Models}},
  author={Rombach, Robin and Blattmann, Andreas and Lorenz, Dominik and Esser, Patrick and Ommer, Bj{\"o}rn},
  booktitle={Proceedings of the IEEE/CVF Conference on Computer Vision and Pattern Recognition},
  pages={10684--10695},
  year={2022}
}

@article{laion400m,
  title={{LAION-400M: Open Dataset of CLIP-Filtered 400 Million Image-Text Pairs}},
  author={Schuhmann, Christoph and Vencu, Richard and Beaumont, Romain and Kaczmarczyk, Robert and Mullis, Clayton and Katta, Aarush and Coombes, Theo and Jitsev, Jenia and Komatsuzaki, Aran},
  journal={arXiv preprint arXiv:2111.02114},
  year={2021}
}

@article{laion5b,
  title={{LAION-5B: An open large-scale dataset for training next generation image-text models}},
  author={Schuhmann, Christoph and Beaumont, Romain and Vencu, Richard and Gordon, Cade and Wightman, Ross and Cherti, Mehdi and Coombes, Theo and Katta, Aarush and Mullis, Clayton and Wortsman, Mitchell and others},
  journal={arXiv preprint arXiv:2210.08402},
  year={2022}
}

@inproceedings{cc12m,
  title={{Conceptual 12M: Pushing Web-Scale Image-Text Pre-Training To Recognize Long-Tail Visual Concepts}},
  author={Changpinyo, Soravit and Sharma, Piyush and Ding, Nan and Soricut, Radu},
  booktitle={Proceedings of the IEEE/CVF Conference on Computer Vision and Pattern Recognition},
  pages={3558--3568},
  year={2021}
}

@inproceedings{controlnet,
  title={{Adding Conditional Control to Text-to-Image Diffusion Models}}, 
  author={Lvmin Zhang and Anyi Rao and Maneesh Agrawala},
  booktitle={IEEE International Conference on Computer Vision},
  year={2023},
}

@inproceedings{gligen,
  title={{GLIGEN: Open-Set Grounded Text-to-Image Generation}},
  author={Li, Yuheng and Liu, Haotian and Wu, Qingyang and Mu, Fangzhou and Yang, Jianwei and Gao, Jianfeng and Li, Chunyuan and Lee, Yong Jae},
  booktitle={Proceedings of the IEEE/CVF Conference on Computer Vision and Pattern Recognition},
  pages={22511--22521},
  year={2023}
}

@article{poole2022dreamfusion,
  title={{DreamFusion: Text-to-3D using 2D Diffusion}},
  author={Poole, Ben and Jain, Ajay and Barron, Jonathan T and Mildenhall, Ben},
  journal={arXiv preprint arXiv:2209.14988},
  year={2022}
}

@inproceedings{SJC,
  title={Score jacobian chaining: Lifting pretrained 2d diffusion models for 3d generation},
  author={Wang, Haochen and Du, Xiaodan and Li, Jiahao and Yeh, Raymond A and Shakhnarovich, Greg},
  booktitle={Proceedings of the IEEE/CVF Conference on Computer Vision and Pattern Recognition},
  pages={12619--12629},
  year={2023}
}

@article{huang2023dreamtime,
  title={DreamTime: An Improved Optimization Strategy for Text-to-3D Content Creation},
  author={Huang, Yukun and Wang, Jianan and Shi, Yukai and Qi, Xianbiao and Zha, Zheng-Jun and Zhang, Lei},
  journal={arXiv preprint arXiv:2306.12422},
  year={2023}
}

@article{chen2023fantasia3d,
  title={Fantasia3d: Disentangling geometry and appearance for high-quality text-to-3d content creation},
  author={Chen, Rui and Chen, Yongwei and Jiao, Ningxin and Jia, Kui},
  journal={arXiv preprint arXiv:2303.13873},
  year={2023}
}

@inproceedings{lin2023magic3d,
  title={{Magic3D: High-Resolution Text-to-3D Content Creation}},
  author={Lin, Chen-Hsuan and Gao, Jun and Tang, Luming and Takikawa, Towaki and Zeng, Xiaohui and Huang, Xun and Kreis, Karsten and Fidler, Sanja and Liu, Ming-Yu and Lin, Tsung-Yi},
  booktitle = {Proceedings of the IEEE/CVF Conference on Computer Vision and Pattern Recognition},
  pages={300-309},
  year={2023}
}

@article{wang2023prolificdreamer,
  title={ProlificDreamer: High-Fidelity and Diverse Text-to-3D Generation with Variational Score Distillation},
  author={Wang, Zhengyi and Lu, Cheng and Wang, Yikai and Bao, Fan and Li, Chongxuan and Su, Hang and Zhu, Jun},
  journal={arXiv preprint arXiv:2305.16213},
  year={2023}
}

@inproceedings{liu2023zero,
  title={Zero-1-to-3: Zero-shot one image to 3d object},
  author={Liu, Ruoshi and Wu, Rundi and Van Hoorick, Basile and Tokmakov, Pavel and Zakharov, Sergey and Vondrick, Carl},
  booktitle={Proceedings of the IEEE/CVF International Conference on Computer Vision},
  pages={9298--9309},
  year={2023}
}

@inproceedings{objaverse,
  title={{Objaverse: A Universe of Annotated 3D Objects}},
  author={Deitke, Matt and Schwenk, Dustin and Salvador, Jordi and Weihs, Luca and Michel, Oscar and VanderBilt, Eli and Schmidt, Ludwig and Ehsani, Kiana and Kembhavi, Aniruddha and Farhadi, Ali},
  booktitle={Proceedings of the IEEE/CVF Conference on Computer Vision and Pattern Recognition},
  pages={13142--13153},
  year={2023}
}

@article{deitke2023objaverse,
  title={Objaverse-xl: A universe of 10m+ 3d objects},
  author={Deitke, Matt and Liu, Ruoshi and Wallingford, Matthew and Ngo, Huong and Michel, Oscar and Kusupati, Aditya and Fan, Alan and Laforte, Christian and Voleti, Vikram and Gadre, Samir Yitzhak and others},
  journal={arXiv preprint arXiv:2307.05663},
  year={2023}
}

@article{liu2023syncdreamer,
  title={SyncDreamer: Generating Multiview-consistent Images from a Single-view Image},
  author={Liu, Yuan and Lin, Cheng and Zeng, Zijiao and Long, Xiaoxiao and Liu, Lingjie and Komura, Taku and Wang, Wenping},
  journal={arXiv preprint arXiv:2309.03453},
  year={2023}
}

@article{weng2023consistent123,
  title={Consistent123: Improve Consistency for One Image to 3D Object Synthesis},
  author={Weng, Haohan and Yang, Tianyu and Wang, Jianan and Li, Yu and Zhang, Tong and Chen, CL and Zhang, Lei},
  journal={arXiv preprint arXiv:2310.08092},
  year={2023}
}

@article{shi2023mvdream,
  title={Mvdream: Multi-view diffusion for 3d generation},
  author={Shi, Yichun and Wang, Peng and Ye, Jianglong and Long, Mai and Li, Kejie and Yang, Xiao},
  journal={arXiv preprint arXiv:2308.16512},
  year={2023}
}

@article{long2023wonder3d,
  title={Wonder3D: Single Image to 3D using Cross-Domain Diffusion},
  author={Long, Xiaoxiao and Guo, Yuan-Chen and Lin, Cheng and Liu, Yuan and Dou, Zhiyang and Liu, Lingjie and Ma, Yuexin and Zhang, Song-Hai and Habermann, Marc and Theobalt, Christian and others},
  journal={arXiv preprint arXiv:2310.15008},
  year={2023}
}

@article{shi2023zero123++,
  title={Zero123++: a Single Image to Consistent Multi-view Diffusion Base Model},
  author={Shi, Ruoxi and Chen, Hansheng and Zhang, Zhuoyang and Liu, Minghua and Xu, Chao and Wei, Xinyue and Chen, Linghao and Zeng, Chong and Su, Hao},
  journal={arXiv preprint arXiv:2310.15110},
  year={2023}
}

@inproceedings{labbe2023megapose,
  title={MegaPose: 6D Pose Estimation of Novel Objects via Render \& Compare},
  author={Labb{\'e}, Yann and Manuelli, Lucas and Mousavian, Arsalan and Tyree, Stephen and Birchfield, Stan and Tremblay, Jonathan and Carpentier, Justin and Aubry, Mathieu and Fox, Dieter and Sivic, Josef},
  booktitle={Conference on Robot Learning},
  pages={715--725},
  year={2023},
  organization={PMLR}
}

@article{zhou2019neurvps,
  title={Neurvps: Neural vanishing point scanning via conic convolution},
  author={Zhou, Yichao and Qi, Haozhi and Huang, Jingwei and Ma, Yi},
  journal={Advances in Neural Information Processing Systems},
  volume={32},
  year={2019}
}

@inproceedings{downs2022google,
  title={Google scanned objects: A high-quality dataset of 3d scanned household items},
  author={Downs, Laura and Francis, Anthony and Koenig, Nate and Kinman, Brandon and Hickman, Ryan and Reymann, Krista and McHugh, Thomas B and Vanhoucke, Vincent},
  booktitle={2022 International Conference on Robotics and Automation (ICRA)},
  pages={2553--2560},
  year={2022},
  organization={IEEE}
}

@article{tremblay2022rtmv,
  title={Rtmv: A ray-traced multi-view synthetic dataset for novel view synthesis},
  author={Tremblay, Jonathan and Meshry, Moustafa and Evans, Alex and Kautz, Jan and Keller, Alexander and Khamis, Sameh and M{\"u}ller, Thomas and Loop, Charles and Morrical, Nathan and Nagano, Koki and others},
  journal={arXiv preprint arXiv:2205.07058},
  year={2022}
}

@Misc{threestudio2023,
  author =       {Yuan-Chen Guo and Ying-Tian Liu and Ruizhi Shao and Christian Laforte and Vikram Voleti and Guan Luo and Chia-Hao Chen and Zi-Xin Zou and Chen Wang and Yan-Pei Cao and Song-Hai Zhang},
  title =        {threestudio: A unified framework for 3D content generation},
  howpublished = {\url{https://github.com/threestudio-project/threestudio}},
  year =         {2023}
}

@article{TOSS,
  title={Toss: High-quality text-guided novel view synthesis from a single image},
  author={Shi, Yukai and Wang, Jianan and Cao, He and Tang, Boshi and Qi, Xianbiao and Yang, Tianyu and Huang, Yukun and Liu, Shilong and Zhang, Lei and Shum, Heung-Yeung},
  journal={arXiv preprint arXiv:2310.10644},
  year={2023}
}

@article{garcia2017review,
  title={A review on deep learning techniques applied to semantic segmentation},
  author={Garcia-Garcia, Alberto and Orts-Escolano, Sergio and Oprea, Sergiu and Villena-Martinez, Victor and Garcia-Rodriguez, Jose},
  journal={arXiv preprint arXiv:1704.06857},
  year={2017}
}

@article{li2023instant3d,
  title={Instant3d: Fast text-to-3d with sparse-view generation and large reconstruction model},
  author={Li, Jiahao and Tan, Hao and Zhang, Kai and Xu, Zexiang and Luan, Fujun and Xu, Yinghao and Hong, Yicong and Sunkavalli, Kalyan and Shakhnarovich, Greg and Bi, Sai},
  journal={arXiv preprint arXiv:2311.06214},
  year={2023}
}

@article{Kingma2014AdamAM,
  title={Adam: A Method for Stochastic Optimization},
  author={Diederik P. Kingma and Jimmy Ba},
  journal={CoRR},
  year={2014},
  volume={abs/1412.6980},
  url={https://doi.org/10.48550/arXiv.1412.6980}
}

@inproceedings{qin2022highly,
  title={Highly accurate dichotomous image segmentation},
  author={Qin, Xuebin and Dai, Hang and Hu, Xiaobin and Fan, Deng-Ping and Shao, Ling and Van Gool, Luc},
  booktitle={European Conference on Computer Vision},
  pages={38--56},
  year={2022},
  organization={Springer}
}

@inproceedings{radford2021learning,
  title={Learning transferable visual models from natural language supervision},
  author={Radford, Alec and Kim, Jong Wook and Hallacy, Chris and Ramesh, Aditya and Goh, Gabriel and Agarwal, Sandhini and Sastry, Girish and Askell, Amanda and Mishkin, Pamela and Clark, Jack and others},
  booktitle={International conference on machine learning},
  pages={8748--8763},
  year={2021},
  organization={PMLR}
}

@article{dosovitskiy2020image,
  title={An image is worth 16x16 words: Transformers for image recognition at scale},
  author={Dosovitskiy, Alexey and Beyer, Lucas and Kolesnikov, Alexander and Weissenborn, Dirk and Zhai, Xiaohua and Unterthiner, Thomas and Dehghani, Mostafa and Minderer, Matthias and Heigold, Georg and Gelly, Sylvain and others},
  journal={arXiv preprint arXiv:2010.11929},
  year={2020}
}

@article{ma2022principles,
  title={On the principles of parsimony and self-consistency for the emergence of intelligence},
  author={Ma, Yi and Tsao, Doris and Shum, Heung-Yeung},
  journal={Frontiers of Information Technology \& Electronic Engineering},
  volume={23},
  number={9},
  pages={1298--1323},
  year={2022},
  publisher={Springer}
}

@article{Mildenhall2020NeRFRS,
  title={NeRF: Representing Scenes as Neural Radiance Fields for View Synthesis},
  author={Ben Mildenhall and Pratul P. Srinivasan and Matthew Tancik and Jonathan T. Barron and Ravi Ramamoorthi and Ren Ng},
  journal={Commun. ACM},
  year={2020},
  volume={65},
  pages={99-106},
  url={https://api.semanticscholar.org/CorpusID:213175590}
}

@article{1987Marching,
  title={Marching cubes: A high resolution 3D surface construction algorithm},
  author={ Lorensen, William E.  and  Cline, Harvey E. },
  journal={ACM SIGGRAPH Computer Graphics},
  pages={163-169},
  year={1987},
}

@article{wang2004image,
  title={Image quality assessment: from error visibility to structural similarity},
  author={Wang, Zhou and Bovik, Alan C and Sheikh, Hamid R and Simoncelli, Eero P},
  journal={IEEE transactions on image processing},
  volume={13},
  number={4},
  pages={600--612},
  year={2004},
  publisher={IEEE}
}

@article{song2020denoising,
  title={Denoising diffusion implicit models},
  author={Song, Jiaming and Meng, Chenlin and Ermon, Stefano},
  journal={arXiv preprint arXiv:2010.02502},
  year={2020}
}
\clearpage
\appendix
\section{More Qualitative Results}
\label{appendix:more_cases}

\subsection{Comparison on consistent multi-view generation.}
\label{appendix:num-rounds-viz}

\begin{figure}[htbp]
  \centering
   \includegraphics[width=0.75\linewidth]{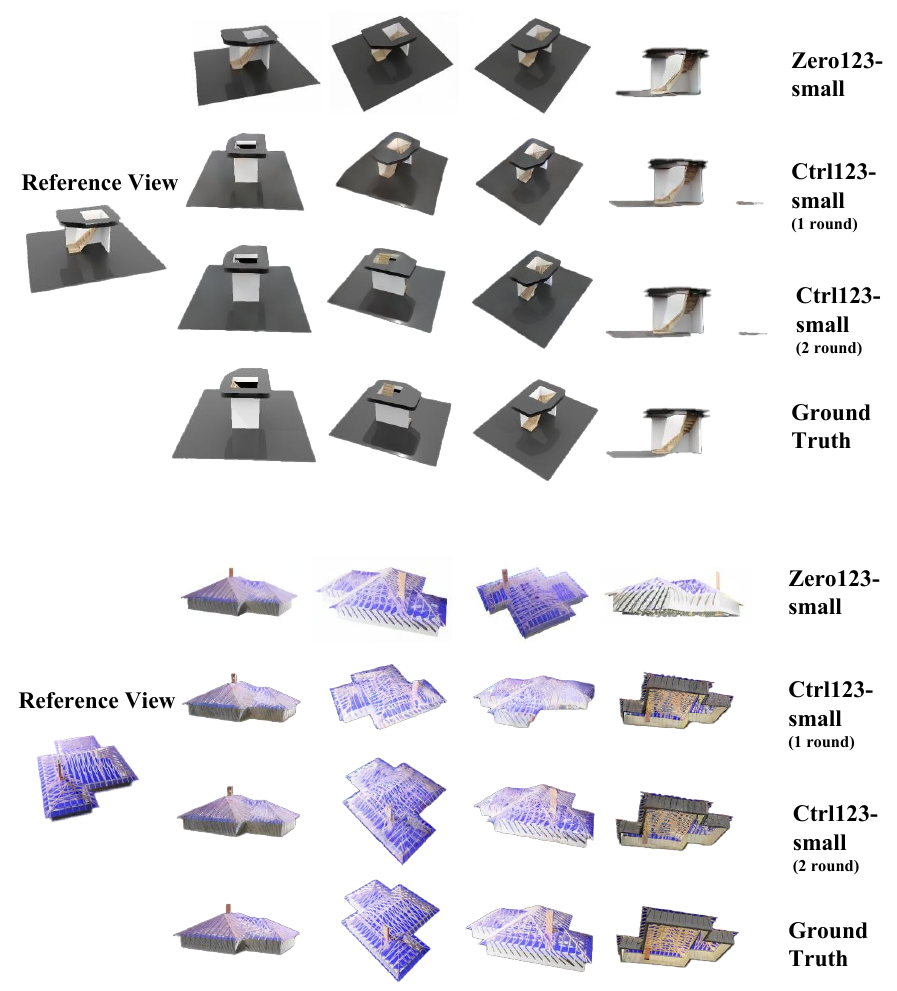}
   \caption{More qualitative comparison of the quality and pose on the generated views with the ground truth views trained on the subset dataset (25 objects). Comparisons between Zero123-small, \ours{}-small, and GT.}
   \label{fig:C1}
\end{figure}

In Figure~\ref{fig:C1}, we showcase more cases comparing the quality and pose consistency between the generated novel views and ground truth. It is evident that \ours{}-small generates much more pose and appearance consistent views for instances in the training set.

\subsection{More results generated by Ctrl123 after training on 100K objects.}
\label{appendix:train_on_100k}

\begin{table*}[htb]
\centering
    \scriptsize
    \setlength{\tabcolsep}{3pt}
    \caption{Quantitative comparison of \ours{} and four baselines, measured with 4 different metrics: PSNR ($\uparrow$), KID ($\downarrow$), AA$^{x^{\circ}}$($\uparrow$), and IoU$^{x}$($\uparrow$), 
     and evaluated on 3 datasets: GSO, RTMV, and 25 randomly sampled objects from the training set.}
    \label{tab:compare_with_zero123_appendix}
    \begin{tabular*}{\hsize}{@{}@{\extracolsep{\fill}}lcccccccccccc@{}}
    \toprule
    \multirow{2}{*}{Method} & \multicolumn{4}{c}{GSO} & \multicolumn{4}{c}{RTMV} & \multicolumn{4}{c}{25 Training Objects} \\
    \cline{2-5} \cline{6-9} \cline{10-13}     
    & PSNR & KID & AA$^{15^{\circ}}$ &IoU$^{0.7}$
    & PSNR & KID & AA$^{15^{\circ}}$ &IoU$^{0.7}$ & PSNR & KID & AA$^{15^{\circ}}$ &IoU$^{0.7}$\\
    \midrule
    Zero123 & 17.2049 & 0.0055 & 14.41\% & 47.32\%&  8.9123&  0.0340& 7.04\% & 2.91\%& 15.9452 & \textbf{0.0043} & 14.28\% & 31.86\%\\
    SyncDreamer \cite{liu2023syncdreamer} & 17.6317 & 0.0077 & 15.94\% & 51.65\% & 9.6344 & 0.0702 & 2.06\% & 1.48\% & / & / & / & /\\
    Zero123++ \cite{shi2023zero123++} & 18.2247 & 0.0140 & 15.63\% & 32.45\% & / & / & / & / & / & / & / & /\\
    Consistent123 \cite{weng2023consistent123} & 13.8337 & 0.0456 & 3.17\% & 29.85\% & 7.1517 & 0.0857 & 1.35\% & 1.06\% & / & / & / & /\\
    \ours{} & \textbf{18.4712} & \textbf{0.0045} & \textbf{16.96\%} & \textbf{62.11\%}& \textbf{10.8938} & \textbf{0.0294} &\textbf{11.98\%} & \textbf{12.41\%}& \textbf{18.9398} & 0.0052 & \textbf{19.79\%} &\textbf{45.97\%} \\
     \bottomrule
    \end{tabular*}%
\end{table*}

\begin{figure}
  \centering
   \includegraphics[width=0.85\linewidth]{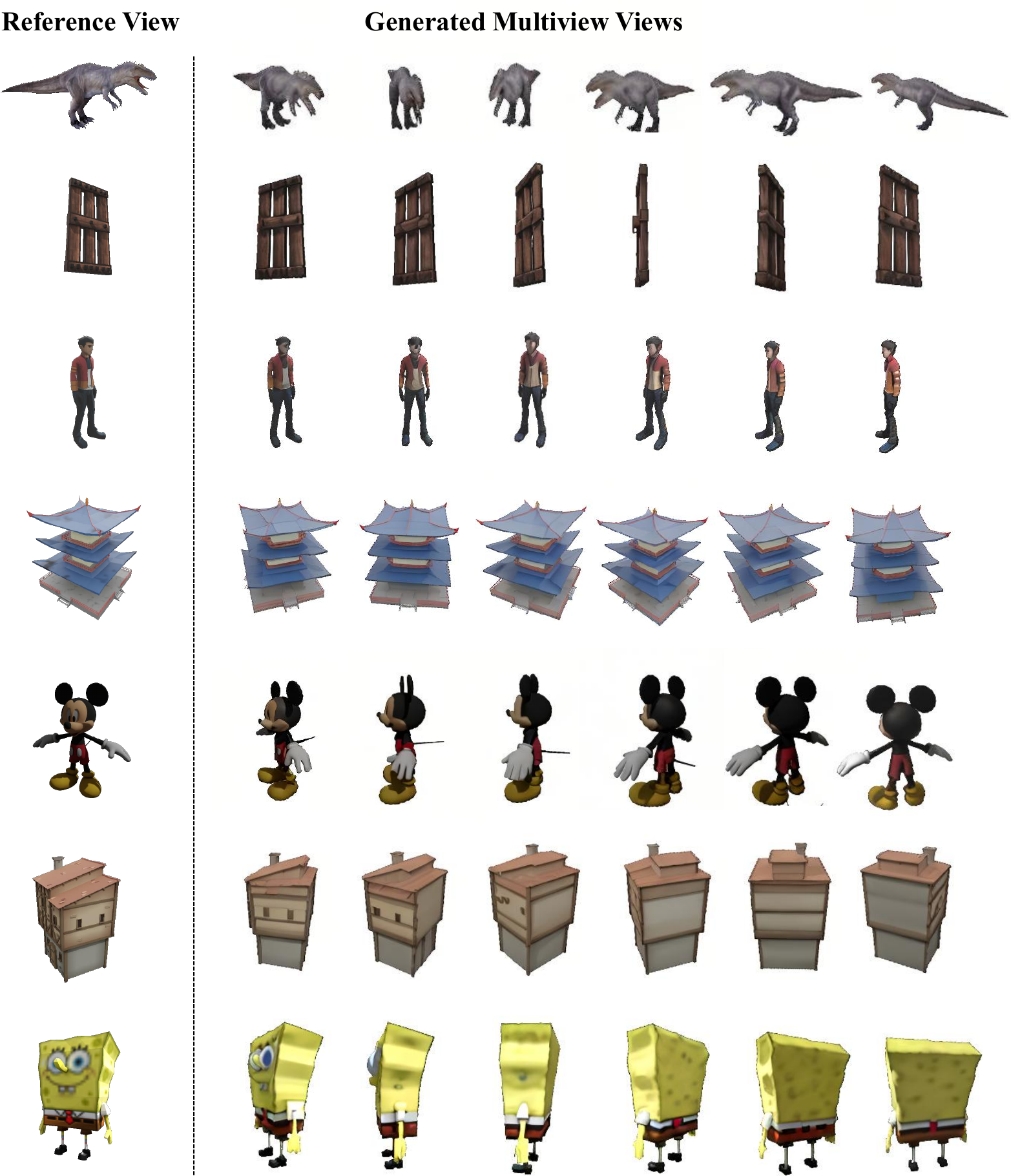}
   \caption{Examples of new views generated by \ours{} for images in  the training dataset (100K Objaverse \cite{objaverse}).}
   \label{fig:C2-1}
   \vspace{-0.7cm}
\end{figure}

\begin{figure}
  \centering
   \includegraphics[width=0.85\linewidth]{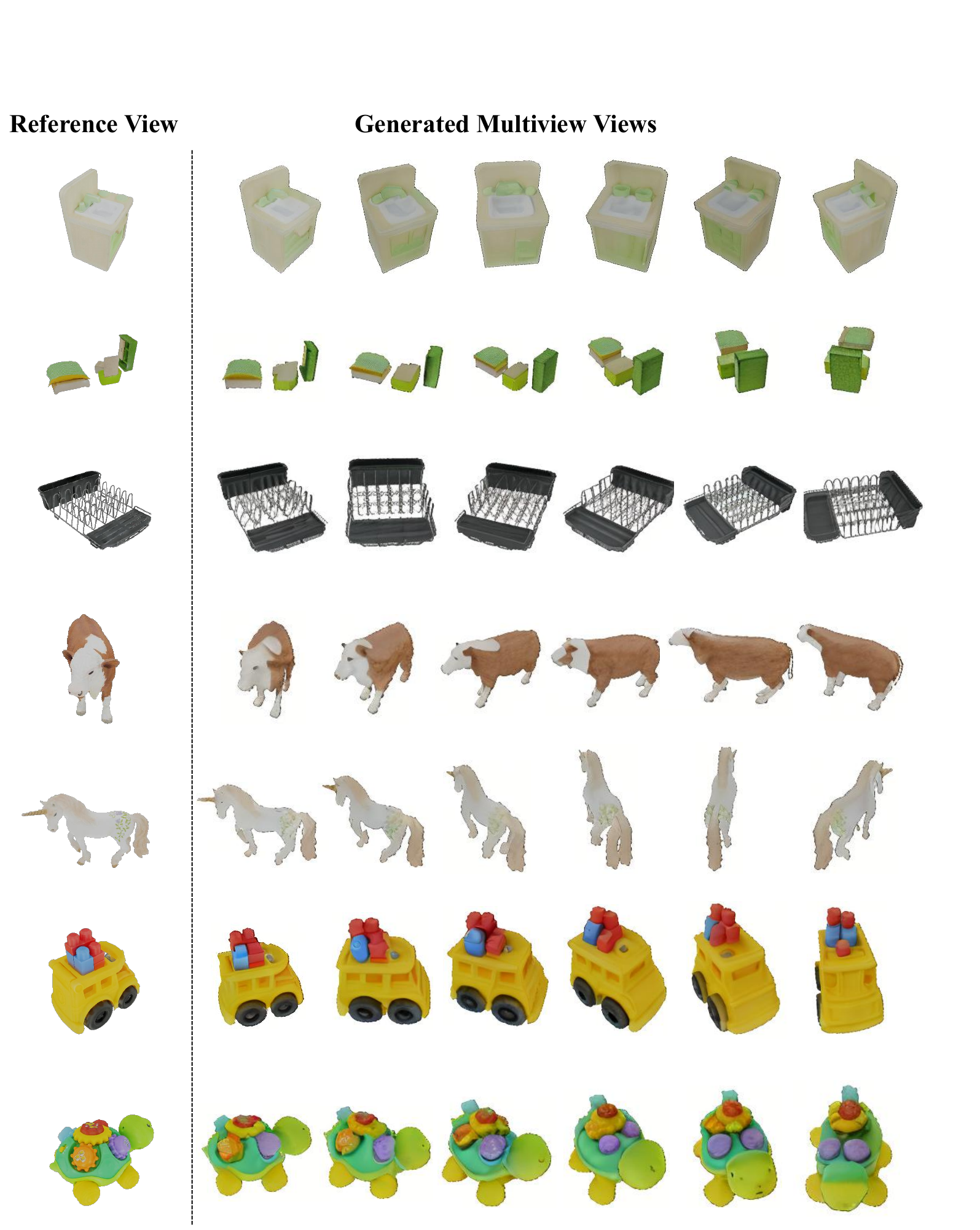}
   \caption{Examples of using \ours{} to generate new views for images in the GSO dataset\cite{downs2022google}.}
   \label{fig:C2-2}
\end{figure}

In this subsection, we present more results of \ours{} after large-scale training on Objectverse \cite{objaverse}. In Figure~\ref{fig:C2-1},  we observe that \ours{} generates high-quality novel views on images from the training dataset; In Figure~\ref{fig:C2-2}, \ours{} also generates high-quality novel views for data outside the training dataset. We also evaluate performance of Zero123++, SyncDreamer, and Consistent123 on RTMV \cite{tremblay2022rtmv} under our settings in Table \ref{tab:compare_with_zero123_appendix}. Note that RTMV \cite{tremblay2022rtmv} only offers images of randomly sampled views, which is not fair for these work when evaluating.

\begin{table}
  \caption{Average angular difference results through Megapose for nearly identical picture pairs.}
  \label{tab:angular_diff}
  \centering
  \begin{tabular}{{llllll}}
    \toprule
    angular difference(through MegaPose)  & view-pair1 &  view-pair2 & view-pair3 & \textit{average} \\ 
    \midrule
    3D\_Dollhouse\_Sink & $3.5833^{\circ}$ & $2.4288^{\circ}$ & $1.0024^{\circ}$ & $5.9619^{\circ}$ \\ 
    \midrule
    JUNGLE\_HEIGHT & $2.0103^{\circ}$ & $6.6258^{\circ}$ & $5.2131^{\circ}$ & $5.0780^{\circ}$ \\ 
    \midrule
    Lenovo\_Yoga\_2\_11 & $3.8219^{\circ}$ & $6.0171^{\circ}$ & $2.9925^{\circ}$ & $6.9773^{\circ}$ \\ 
    \midrule
    Shark & $5.2212^{\circ}$ & $3.9884^{\circ}$ & $6.0070^{\circ}$ & $5.2961^{\circ}$ \\
    \midrule
    Sonny\_School\_Bus & $6.9927^{\circ}$ & $4.5645^{\circ}$ & $15.1978^{\circ}$ & $7.4337^{\circ}$ \\
    \bottomrule
  \end{tabular}
\end{table}

\begin{table}[htb]
\vspace{-0.7cm}
  \caption{AA values for different thresholds of multi-rounds Ctrl123 when trained on a subset of the curated datase.}
  \label{tab:diff_thre}
  \centering
  \begin{tabular}{{llllll}}
    \toprule
    Method  & AA$^{5^{\circ}}$ & AA$^{10^{\circ}}$ & AA$^{15^{\circ}}$ & AA$^{20^{\circ}}$ & \textit{AA(Average)}\\ 
    \midrule
    Zero123-small & 6.29\% & 12.55\% & 22.62\% & 32.20\% & 18.42\%\\
    \midrule
    \ours{}-small & \multirow{2}{*}{14.23\%} & \multirow{2}{*}{21.25\%} & \multirow{2}{*}{32.02\%} & \multirow{2}{*}{41.89\%} & \multirow{2}{*}{27.35\%} \\
    (1 round)   & & & & & \\
    \midrule
    \ours{}-small & \multirow{2}{*}{30.41\%} & \multirow{2}{*}{39.34\%} & \multirow{2}{*}{57.78\%} & \multirow{2}{*}{69.51\%} & \multirow{2}{*}{49.26\%} \\
    (2 round)   & & & & & \\
    \bottomrule
  \end{tabular}
\end{table}

\section{The credibility of our introduced AA metric}
\label{sec: AA credibility}
\subsection{Precise single image pose estimation by MegaPose}

To evaluate the accuracy of pose estimation through Megapose \cite{labbe2023megapose}, we randomly select 5 objects from GSO dataset \cite{downs2022google} and randomly sampled 12 view pairs, ensuring that the angular difference between each pair is less than 1° but not equal to 0 (to avoid identical view pairs). Subsequently, we utilized MegaPose \cite{labbe2023megapose} to estimate the average angular difference for each object. For brevity, Table \ref{tab:angular_diff} only displays results for three cases and the average results for 12 cases. In Table \ref{tab:angular_diff}, Only one pair (view-pair3 of Sonny\_School\_Bus) results in an error exceeding 15°, while the average error for each object is less than 7.5° (half of 15°). These findings prove the credibility of MegaPose \cite{labbe2023megapose} and our setting for the predefined threshold 15° of our introduced AA metric.

\subsection{Our setting for the predefined threshold of our proposed AA metric}

One may concern that AA of Ctrl123 is only better than that of Zero123 under a certain threshold. To eliminate this type of concern, we select 4 thresholds (5, 10, 15, and 20) and computed the corresponding AA values. Table \ref{tab:diff_thre} shows results for the experiment in Paragraph \textit{Improvement with Multi-rounds Optimization} of Section \ref{sec:exp:largescale}. Results in Table \ref{tab:diff_thre} proves the credibility of AA when the threshold is set as $15^{\circ}$.

\begin{table}[htb]
  \caption{Ablation study on the number of denoise steps for $\hat{\X}_{tg}$ generation.}
  \label{tab:ablation_denoise_step}
  \centering
  \begin{tabular}{{lllllll}}
    \toprule
    Method vs.  & \multirow{2}{*}{Image Quality} &  \multirow{2}{*}{Alignment(MegaPose/DIS)} \\ 
    Metrics     &     &    \\  
    \midrule
    \multirow{2}{*}{Zero123} & KID :0.0254 & AA$^{15^{\circ}}$:22.62\% \\
                              & PSNR :19.3839 & IoU$^{0.7}$:30.93\% \\
    \midrule
    \ours{} & KID:0.0194 & AA$^{15^{\circ}}$:23.56\% \\
    (1 denoise steps)    & PSNR :20.1593 &  IoU$^{0.7}$:32.18\% \\
    \midrule
    \ours{} & KID:0.0102 & AA$^{15^{\circ}}$:25.71\% \\
    (10 denoise steps)    & PSNR:22.5739 &  IoU$^{0.7}$:38.97\% \\
    \midrule
    \ours{} & KID:0.0095 & AA$^{15^{\circ}}$:29.47\% \\
    (30 denoise steps)    & PSNR:23.1165 & IoU$^{0.7}$:49.32\%  \\
    \midrule
    \ours{} & KID: 0.0087 & AA$^{15^{\circ}}$:32.02\%\\
    (50 denoise steps)    & PSNR:23.6080 &  IoU$^{0.7}$: 55.01\%  \\
    \bottomrule
  \end{tabular}
\end{table}

\section{Implementation Details for Image-to-3D Generation}
\label{appendix:image23d_imp_details}

\begin{figure}[htb]
  \centering
   \includegraphics[width=0.7\linewidth]{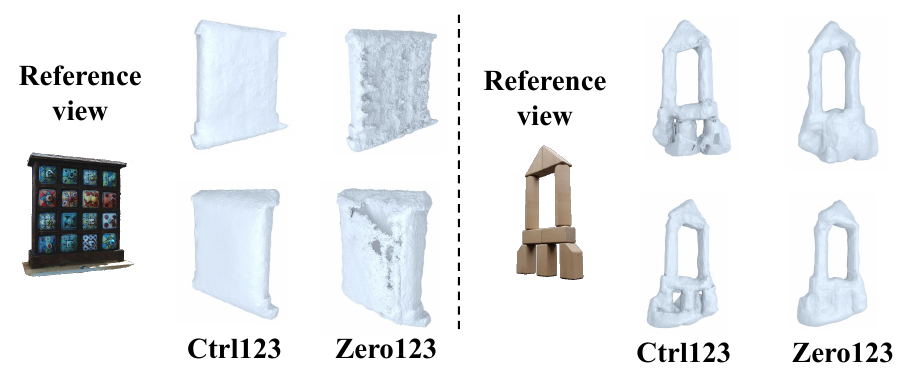}
   \caption{Qualitative comparison of 3D reconstruction from single view images with different methods (\ours{} vs Zero123).}
   \label{fig:comparison_on_3D_nvs}
   \vspace{-0.7cm}
\end{figure}

In Section~\ref{sec:exp:3D}, we demonstrate the enhancement of 3D reconstruction quality by substituting Zero123 with \ours{}.  All results are obtained using the SDS \cite{poole2022dreamfusion} loss implemented by threestudio \cite{threestudio2023}. In threestudio, we input one image and train a NeRF\cite{Mildenhall2020NeRFRS} with 800 training steps. For each training step, we set the noise scale from 0.4 to 0.85. We employ the Adam optimizer\cite{Kingma2014AdamAM} with $lr=0.01$, $\beta_1=0.9$ and $\beta_2=0.99$. After training, in order to get more detailed models, we use marching cube method \cite{1987Marching} to export NeRF to mesh.

\section{More Ablation Study}
\label{appendix:ablation}
\begin{table}[htb]
  \caption{Ablation study on the optimization strategies.}
  \label{tab:ablation_opt_strategy}
  \centering
  \begin{tabular}{{lll}}
    \toprule
    Strategy  & Image Quality &  Alignment(MegaPose/DIS) \\ 
    \midrule
    \multirow{2}{*}{Simultaneous} & KID:0.0218 & AA$^{15^{\circ}}$:25.21\% \\
      & PSNR:17.9643 &  IoU$^{0.7}$:35.41\% \\
    \midrule
    \multirow{2}{*}{Alternative} & KID:0.0087 & AA$^{15^{\circ}}$:32.02\%  \\
    & PSNR:23.6080 &  IoU$^{0.7}$: 55.01\%  \\
    \bottomrule
  \end{tabular}
\end{table}
\paragraph{Ablation study on denoising steps when CL training.} It is slow to generate a sample with DDPM \cite{ddpm} by following the Markov chain of the reverse diffusion process, as $t_\infty$ can as many as a few thousand steps. One simple way to accelerate the process is to run a strided sampling schedule (DDIM) \cite{song2020denoising} with every $\lceil t_\infty/n \rceil$ steps to reduce the process from $t_\infty$ to $n$ steps. The results, presented in Table~\ref{tab:ablation_denoise_step}, indicate that the pose alignment improves as the number of denoising steps increases while image quality improvement stops at the 30 steps. We choose denoising step 50 since the pose alignment is the property that we care the most.

\paragraph{Simultaneous vs Alternative training strategy.} In Table~\ref{tab:ablation_opt_strategy}, the row ``Alternative'' is the results of the row ``\ours{}-small(1 round)'' in Table~\ref{tab:ablation_rounds_number}. We implement ``Simultaneous'' by changing the loss of the experiment ``\ours{}-small(1 round)'' from the alternative version to the simultaneous version. All other settings like learning rate, optimizer etc are the same. The results in Table~\ref{tab:ablation_opt_strategy} shows the ``Alternative'' strategy is better. Note that we also try adjusting related hyper-parameters to increase the performance of the simultaneous version training. However, all results show bad performance.

\begin{table}[htb]
\centering
    \small
    \caption{Ablation study on the number of rounds for alternative training strategy when trained on a subset of the curated dataset, measured with 4 metrics: KID($\downarrow$), PSNR($\uparrow$), AA$^{x^{\circ}}$($\uparrow$), and IoU$^{x}$($\uparrow$).}
    \label{tab:zero123 vs small}
    \begin{tabular*}{\hsize}{@{}@{\extracolsep{\fill}}llllllll@{}}
    \toprule
    Method vs.  & Training & \multirow{2}{*}{Image Quality} &  Alignment  \\ 
    Metrics     & steps  &  & (MegaPose/DIS)  \\ 
    \midrule

    \multirow{2}{*}{Zero123-small} & \multirow{2}{*}{20,000} & KID:0.0254 & AA$^{15^{\circ}}$:22.62\% \\
                            &  & PSNR:19.3839 & IoU$^{0.7}$ :30.93\% &  \\
    \midrule
    \multirow{2}{*}{Zero123} & \multirow{2}{*}{10,000} & KID: 0.0041& AA$^{15^{\circ}}$:26.40\% \\
    & & PSNR:19.3163 &  IoU$^{0.7}$ :55.35\% \\
    \bottomrule
    \end{tabular*}%
    \vspace{-0.4cm}
\end{table}


\paragraph{Zero123 vs Zero123-small.} As shown in Table \ref{tab:zero123 vs small}, we evaluate the performance of Zero123 and Zero123-small on the same 25 objects. Based on that Zero123-small has already converged, Zero123-small shows worse pose-inconsistency (based on the comparison of metrics: AA$^{x^{\circ}}$ and IoU$^{x}$).

\begin{figure}[htb]
   \centering
   \includegraphics[width=0.5\linewidth]{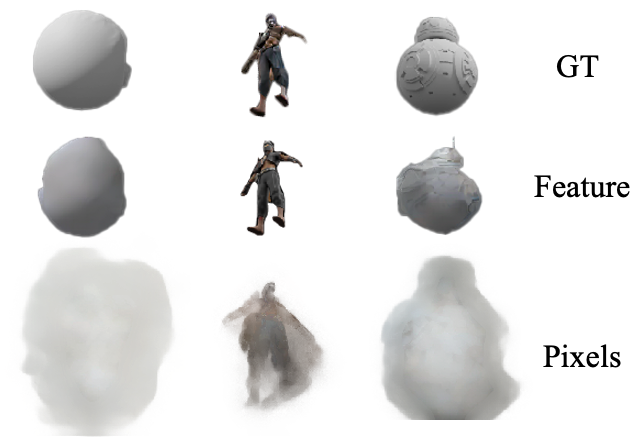}
   \caption{Qualitative comparison on the difference type consistency loss (Pixel space vs Latent/feature space).}
   \label{fig:ablation-pixel-vs-feat}
   \vspace{-0.4cm}
\end{figure}

\begin{table}[H]
  \caption{Ablation study on different types of consistency losses (pixel space vs latent/feature space).}
  \label{tab:ablation_consistency_type}
  \centering
  \begin{tabular}{{lllllll}}
    \toprule
    Consistency  & \multirow{2}{*}{Image Quality} &  \multirow{2}{*}{Alignment(MegaPose/DIS)} \\ 
    Type     &     &    \\
    \midrule

    \multirow{2}{*}{Pixels} & KID:0.1490 & AA$^{15^{\circ}}$:4.61\% \\
                              & PSNR:10.6905 & IoU$^{0.7}$:21.49\% \\
    \midrule
    \multirow{2}{*}{Feature} & KID:0.0301 & AA$^{15^{\circ}}$:29.75\% \\
                              & PSNR:18.7103 & IoU$^{0.7}$:48.19\% \\
    
    \bottomrule
  \end{tabular}
\end{table}

\begin{table}[H]
  \caption{Comparsion of the closed-loop loss on class feature vs patch feature.}
  \label{tab:ablation_feature_type}
  \centering
  \begin{tabular}{{lllllll}}
    \toprule
    Feature  & \multirow{2}{*}{Image Quality} &  \multirow{2}{*}{Alignment(MegaPose/DIS)} \\ 
    Type     &     &    \\ 
    \midrule

    \multirow{2}{*}{Class Features} & KID:0.0677 & AA$^{15^{\circ}}$:18.40\% \\
                              & PSNR:14.5749 & IoU$^{0.7}$:32.44\% \\
    \midrule
    \multirow{2}{*}{Patch Features} & KID:0.0301 & AA$^{15^{\circ}}$:29.75\% \\
                              & PSNR:18.7103 & IoU$^{0.7}$:48.19\% \\
    
    \bottomrule
  \end{tabular}
\end{table}

\paragraph{Pixel space vs Latent space.} To verify the effectiveness of the closed-loop framework, we conducted an ablation study that applies mean square error (MSE) directly between $\hat{\X}_{tg}$ and $\X_{tg}$. As shown in Table~\ref{tab:ablation_consistency_type} and Figure~\ref{fig:ablation-pixel-vs-feat}, enforcing MSE directly in pixel space does not lead to the expected convergence and results in divergence. This highlights the challenges in achieving consistency in different representation spaces.

\paragraph{Closed-loop loss on class features vs patch features.} As discussed in section~\ref{sec:method:ctrl123}, the choice of closed-loop loss on the class feature $\bm Z_{tg,c}$ or the patch feature $\bm Z_{tg,p}$ is important for the method. Hence, we conduct the ablation study to experiment with the different choices of class features and patch features. As shown in Table \ref{tab:ablation_feature_type}, loss on the patch features could improve the image quality and alignment better than that on the class feature.

\begin{table}[H]
  \caption{Comparsion of Zero123 performance with the different code base: threestudio \cite{threestudio2023} vs zero123\cite{liu2023zero}. The evaluation setting follows Table \ref{tab:compare_with_zero123}}
  \label{tab:code base}
  \centering
  \begin{tabular}{{lll}}
    \toprule
    Code Base & Zero123 & threestudio \cite{threestudio2023} \\
    \midrule
    PSNR & 18.1865 & 17.1495 \\
    \bottomrule
  \end{tabular}
  \vspace{-0.7cm}
\end{table}

\begin{table}[H]
  \caption{Comparsion of Ctrl123 performance with the different random seeds. The evaluation setting follows Table \ref{tab:compare_with_zero123}}
  \label{tab:seed}
  \centering
  \begin{tabular}{{llllll}}
    \toprule
    random seed & 35 & 40 & 45 & 50 & 55 \\
    \midrule
    PSNR & 18.4531 & 18.4767 & 18.4712 & 18.3951 & 18.4106\\
    \bottomrule
  \end{tabular}
\end{table}

\paragraph{Different code base.} In our work, we use threestudio \cite{threestudio2023} to conduct all experiments. Note that threestudio uses a more efficient but worse quality inference method, which leads to lower metrics. Table \ref{tab:code base} presents the image quality of generated novel views of Zero123 with different code base.

\paragraph{Errors on different random seeds.} Note that different random seeds will cause the difference of $\hat{\X}_{t_\infty}$, which will lead to the variance of generated novel views. Table \ref{tab:seed} illustrates PSNR of novel views generated by Ctrl123 with different random seeds. Results presented in Table \ref{tab:seed} prove that random seed doesn't influence our claim of Ctrl123.

\end{document}